\documentclass[letterpaper]{article}
\usepackage{confstyle}
\usepackage[hyphens]{url}
\usepackage{graphicx}
\usepackage[round]{natbib} 
\usepackage{caption}
\usepackage{algorithm}
\usepackage{algorithmic}

\usepackage{newfloat}
\usepackage{listings}
\DeclareCaptionStyle{ruled}{labelfont=normalfont,labelsep=colon,strut=off}
\floatstyle{ruled}
\newfloat{listing}{tb}{lst}{}
\floatname{listing}{Listing}

\usepackage{booktabs}

\usepackage[utf8]{inputenc}
\usepackage[T1]{fontenc}
\usepackage{url}
\usepackage{amsfonts}
\usepackage{nicefrac}
\usepackage{microtype}
\usepackage{xcolor}

\usepackage{amsmath}
\usepackage{amssymb}
\usepackage{mathtools}
\usepackage{amsthm}

\usepackage{tikz}
\usetikzlibrary{positioning, arrows.meta, calc, fit, backgrounds}
\definecolor{specfill}{RGB}{237,243,251}
\definecolor{specborder}{RGB}{96,128,176}
\definecolor{latentfill}{RGB}{214,228,245}

\usepackage{multirow}
\usepackage{bm}

\newcommand{\edpmethod}{{NEXT}}
\newcommand{\pdemethod}{\edpmethod{}}
{}

\theoremstyle{plain}

\theoremstyle{definition}

\theoremstyle{remark}

\title{
NEXT: Physics-Informed Neuro-Spectral Exponential Time Differencing Architectures
}
\author{
Márcio Marques\textsuperscript{\rm 1}\textsuperscript{*},
Leonardo Mendonça\textsuperscript{\rm 1}\textsuperscript{*},
Leonardo M. Moreira\textsuperscript{\rm 2}\textsuperscript{*},
Christian Júnior de Oliveira\textsuperscript{\rm 1},
Vitor Balestro\textsuperscript{\rm 3},
Tiago Novello\textsuperscript{\rm 1},
Daniel Yukimura\textsuperscript{\rm 1},
Pavel Petrov\textsuperscript{\rm 1},
Lucas Nissenbaum\textsuperscript{\rm 1}
}
\affiliations{
\textsuperscript{\rm 1}Instituto de Matemática Pura e Aplicada (IMPA), Rio de Janeiro, Brasil\\
\textsuperscript{\rm 2}Universidade do Estado do Rio de Janeiro (UERJ), Rio de Janeiro, Brasil\\
\textsuperscript{\rm 3}Universidade Federal Fluminense (UFF), Niterói, Brasil\\
Corresponding: \texttt{marcio.marques@impa.br}
}

\begin{document}

\maketitle
\footnotetext[1]{These authors contributed equally to this work.}

\begin{abstract}
Physics-Informed Neural Networks (PINNs) build neural representations of time-dependent PDE solutions, naturally incorporating physics knowledge and observational data, which makes them well suited to both forward and inverse PDE problems. PINNs, however, are known to suffer from spectral bias and lack of causality. Neuro-Spectral Architectures (NeuSA), a recently proposed alternative to PINNs, mitigate both issues, but their numerical integration becomes unstable for stiff differential equations arising in many relevant physical problems. This study proposes Neuro-Spectral Exponential Time Differencing Architectures (NEXT), which combines the spectral representation of the PDE solution in NeuSA with high-order exponential integrators. Within this approach, the linear stiff part of the vector field induced by the PDE is integrated exactly through matrix exponentials, while the possibly nonlinear remainder is modeled by a neural network. The effectiveness of NEXT is verified through benchmark experiments on a set of stiff PDEs, in which NEXT is stable and accurate while NeuSA diverges numerically. It is also shown that NEXT can be applied to inverse problems, where the model has to learn unknown parameters or boundary conditions from sparse data. All code used in this work is publicly available at: \texttt{https://github.com/marcioh2m/next.git}.

\end{abstract}

\section{Introduction}

Physics-informed neural networks (PINNs)~\cite{raissi2019physics} are a scientific machine learning paradigm where a neural network is trained to represent the solution of a partial differential equation (PDE). 
Given a PDE boundary-value problem 
\begin{equation}
\begin{aligned}
    \mathcal{R}(u)(\mathbf{x},t) =& 0 \,, \quad (\mathbf{x}, t) \in \Omega \,, \\
    \mathcal{B}(u)(\mathbf{x}, t) =& 0 \,, \quad (\mathbf{x}, t) \in \partial\Omega \,,
\end{aligned}
\end{equation}
where $\mathcal{R}$ and $\mathcal{B}$ are (possibly nonlinear) differential operators defined on the domain $\Omega$ and its boundary, respectively, PINNs train a neural surrogate of $u(\mathbf{x},t)$ by minimizing a \textit{physics-informed loss}:
\begin{equation}
    \mathcal{L} = \sum_{(\mathbf{x}_i, t_i) \in \Omega} (\mathcal{R}(u)(\mathbf{x}_i,t_i))^2
    + \sum_{(\mathbf{x}_j, t_j) \in \partial\Omega} (\mathcal{B}(u)(\mathbf{x}_j,t_j))^2 \,,
\label{eq:pinn_loss}
\end{equation}
and the derivatives required for evaluating $\mathcal{R}$ and $\mathcal{B}$ are computed through automatic differentiation.

Compared to classical numerical methods, PINNs enjoy the flexibility of producing an implicit representation that can incorporate data easily as an additional loss term \cite{wu2023-rad, wu2024ropinn}, and use their loss-based formulation to solve inverse problems \citep{hou2024kinematicwave,shang2025simultaneous}.
Due to these advantages, physics-informed neural networks have been applied to a variety of problems, in contexts spanning fluid dynamics~\cite{donnelly2024hydro,molina2024modeling}
, seismic and acoustic wave propagation \cite{ding2024papermarcinho,marques2025,mendonca2026physicsinformed} and other areas of science and engineering \cite{de2022weak,patel2022thermodynamically}. 

The capabilities of PINNs in many practical problems, however, are substantially restricted by two major problems: \textit{spectral bias}, due to multilayer perceptrons (MLPs) inherent difficulty in learning high-frequency components of functions \cite{wang2022-pinnsfail,xu2019frequencyprinciple}; and the \textit{lack of causality}, when the PINN may fail to enforce or propagate initial conditions of a PDE \cite{wang2024causality, yu2022gradient}.

In order to mitigate these issues, Neuro-Spectral Architectures (NeuSA)~\cite{Bizzi2025NeuroSpectral} propose to use spectral-domain representations to solve a PDE by converting it to a system of ordinary differential equations (ODEs), and using a Neural ODE ~\cite{chen2019neural} to solve these differential equations. 

While this approach improves representational fidelity for PDEs, its results are limited to a small set of differential equations. In particular, the sequential nature of time integration makes training over long horizons computationally expensive and renders Neural ODE–based methods vulnerable to stiffness, which can cause NeuSA to diverge unless the integration timestep is extremely small \citep{chandra2026oscillatory}. 

In order to resolve these issues, we propose to incorporate Exponential Time Differencing (ETD) into Neuro-Spectral Architectures. This leads to Neuro-spectral EXponential Time-differencing architectures (\edpmethod{}) that combine the representation power of Neuro-Spectral Architectures with the robustness and computational efficiency of high-order ETD methods when representing nonlinear dynamics. Our contribution is thus an improvement to the state of the art:

\begin{itemize}
\item By using Neuro-Spectral Architectures' decomposition and propagation, {\edpmethod{}} inherits the capability to represent high-frequency components and is causal by construction.
\item By adopting an exponential integrator, \edpmethod{} becomes better suited to stiff dynamics, significantly expanding the variety of PDEs to which the method can be applied, and achieving significant improvements over the baselines in stiff PDEs presented in our experiments, such as the Heat equation as seen in Fig. 1. 
\item Improved stability and computational efficiency of \pdemethod{} make it suitable for tackling parameter identification problems for the PDEs, including cases where NeuSA fails due to the stiffness of the ODE system obtained after a spectral discretization.
\end{itemize}

\begin{figure}[ht]
\centering
\includegraphics[width=\linewidth]{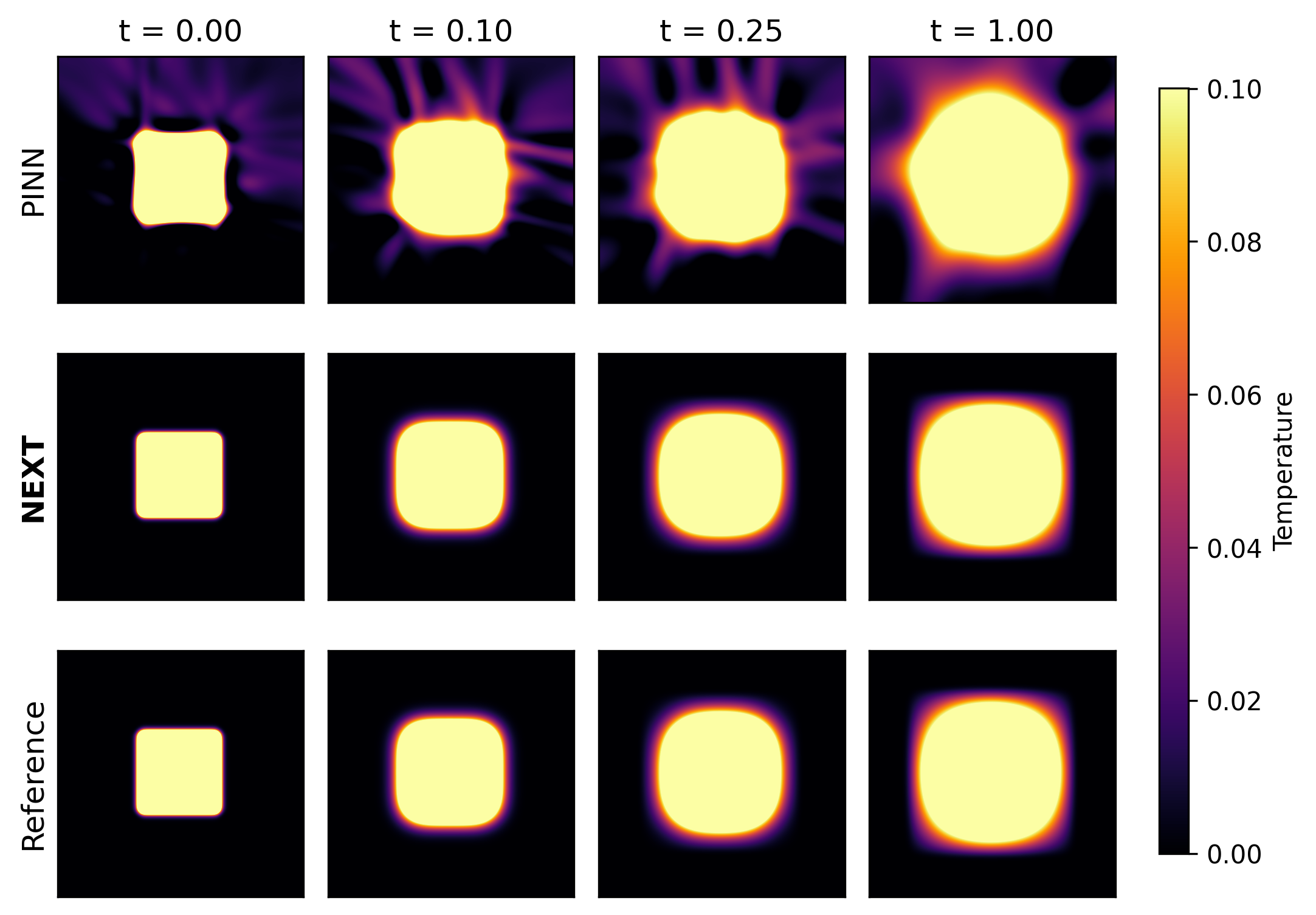}
\caption{\pdemethod{} combines the spectral fidelity and causality of NeuSA with the robustness to stiffness of exponential integrators. This makes it particularly suitable for stiff time-dependent PDEs with sharp interfaces, such as the heterogenous heat equation shown above.}
\label{fig:teaser}
\end{figure}

\section{Neuro-Spectral Architectures}

\textbf{Neural ODEs} are ODEs whose vector field is represented by a neural network. Their training relies on applying numerical solvers to the learned vector fields, comparing generated trajectories with existing solution data, and optimizing the network parameters by gradient descent. The inference of a neural ODE model is also accomplished via the integration of the optimized vector field by a suitable numerical method (usually, a classical explicit Runge-Kutta scheme). 
The Neural ODE technique is thus a hybrid of machine learning and numerical analysis~\cite{dupont2019augmented,kidger2022neural}. 
Being highly efficient across a range of important applications, Neural ODEs are currently widely used in areas such as medical image registration~\cite{balakrishnan2019voxelmorph, wu2022nodeo, sun2022topology, sun2024medical}, chemistry~\cite{Owoyele2022}, computer graphics~\cite{novello2025}, and many others. While many other ODE/PDE solvers powered by neural networks struggle to maintain causality, it is enforced by construction in the Neural ODE framework.

Recently, neural ODEs in combination with spectral discretizations have been successfully applied to the solution of PDEs within the framework of \textbf{physics-informed neuro-spectral architectures  (NeuSA)}~\cite{Bizzi2025NeuroSpectral,loya2025}. 
In these approaches, the solution is represented in a spectral basis (e.g., Fourier modes), reducing the PDE to a system of ODEs in coefficient space, and a neural network is adopted as an implicit neural representation of the governing vector field of this system.

By combining the flow logic of Neural ODEs with the differential capabilities of pseudo-spectral methods, NeuSA becomes a structure that matches time flow, deals well with \textit{spectral bias} and \textit{lack of causality} and is versatile enough to operate in partial differential equations.
This hybrid formulation has managed to achieve high accuracy and fast training in benchmark problems where MLP-based PINNs are known to struggle \citep{Bizzi2025NeuroSpectral, balestro2026neurospectral}.

Consider $d$-dimensional evolutionary partial differential equations of the form
\begin{subequations}
\label{eq:time_PDE}
\begin{align}\label{EDP}
&u_t = \mathcal{F}(u) \,, \quad \mathbf{x} \in \Omega, \quad t \in [0,T]\,,\\
&u(\mathbf{x},0) = g(\mathbf{x})\,, \quad \mathbf{x} \in \Omega\,,\\
&u(\mathbf{x},t) = h(\mathbf{x},t)\,,  \quad \mathbf{x} \in \partial\Omega\,, \quad t \in [0,T]\,,
\end{align}   
\end{subequations}

where $\mathcal{F}$ is a differential operator acting on $u$. The pseudo-spectral decomposition creates a representation of the PDE solution as
\begin{equation}
    u(t,\mathbf{x}) = \sum^{N_0}_{n_0=0}...\sum^{N_d}_{n_d=0}\hat{\mathbf{u}}_{n_0...n_{d}}(t) \prod^{d}_{i=0} b^i_{n_i}(x_i),
\end{equation}
where $b^i_{n_i}$ are spectral basis functions (e.g. sines, cosines or complex exponentials).
Via orthogonal projection onto $b^i_{n_i}$, one can write eq.~\eqref{eq:time_PDE} as an ODE system of the form
\begin{equation}\label{associatedode}
\begin{aligned}
\mathbf{\hat{u}}_t = \mathbf{F}(\mathbf{\hat{u}})\,,\\
\mathbf{\hat{u}}(0) = \mathbf{\hat{u}}_0\,,
\end{aligned}
\end{equation}
where $\mathbf{F}$ is a vector field given by the projection of $\mathcal{F}$ on the finite basis, constructed by evaluating the derivatives of each basis function $b_{n_i}^i$ on the grid points $x_i$ \citep{trefethen2000spectral}. 
A suitable choice of basis can also enforce boundary conditions (e.g. Fourier basis and periodic conditions), in which case eq.~\eqref{associatedode} is equivalent to eq.~\eqref{eq:time_PDE}.

NeuSA proposes to use a neural representation $\mathbf{F}_\theta$ for the vector field $\mathbf{F}$ and learn the parameters $\theta$ through a physics-informed loss as in  \eqref{eq:pinn_loss}, in which the residues are evaluated over the integrated trajectories $\hat{\mathbf{u}}_\theta(t)$. 
In problems where the linear part $\mathbf{L}$  of $\mathbf{F}$ can be naturally separated it was found advantageous to use a vector field given by  $ \tilde{\mathbf{F}}_\theta = \mathbf{L} + \mathbf{F}_\theta $. Although this formulation succeeds when applied to certain benchmark PDEs, 
NeuSA inherits important limitations from both spectral methods and Neural ODEs \citep{Bizzi2025NeuroSpectral}. First, the approach remains strongly dependent on the choice of numerical integrator used to evolve the learned dynamics. Second, the sequential nature of time integration introduces significant computational overhead for large values of $T$. 

Most importantly, spectral discretizations of PDEs (especially those involving diffusive or dispersive operators) often lead to \textbf{stiff systems}. A dynamical system is said to be stiff when it contains multiple time scales, typically characterized by rapidly decaying modes coexisting with slow dynamics, which impose severe stability constraints on explicit time-stepping methods. 
For instance, in the Fourier-basis discretization of PDEs with high-order derivatives in space, such as Korteweg--De Vries (eq.~\eqref{eq:kdv}) 
, the components of $\mathbf{L}$ associated with high-frequency modes are orders of magnitude larger than those of the low-frequency modes \citep{fornberg1999nonlinearwave}. 
As a result, standard explicit integrators require prohibitively small time steps to maintain stability, making integration inefficient and training unfeasible~\cite{hochbruck2010exponential,leveque2007finite,canuto2007spectral,fronk2025training}. For more information on stiffness, see Appendix A.1. In the following section, we propose to combine NeuSA with exponential integrators, to overcome the difficulties associated with such stiffness.

\section{\pdemethod{} architecture}

The stiffness that limits NeuSA is precisely the obstacle that motivated the development of \textbf{exponential time-differencing} (ETD) schemes in the numerical-analysis community. 
ETD methods treat stiff linear dynamics analytically via matrix exponentials, while handling nonlinear terms numerically~\cite{cox2002exponential,kassam2005etd,hochbruck2010exponential}. By directly addressing the primary source of stiffness, ETD methods provide improved stability compared to classical Runge–Kutta or multistep schemes and allow for significantly larger time steps.

\begin{figure}[ht]
    \centering
    \includegraphics[height=1.3\linewidth]{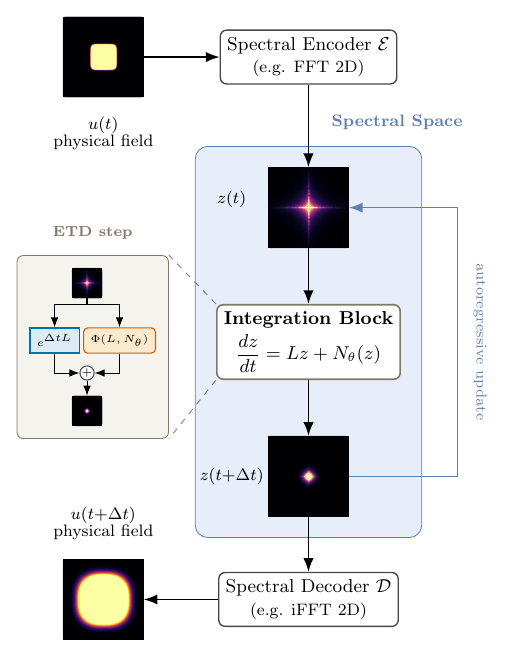}
                \caption{\edpmethod{} consists of a recurrent architecture in which the state $\textbf{u}$ is decomposed into basis spectral components~$\hat{\textbf{u}}_n$. Each component evolves according to an ODE whose linear part is integrated exactly via the matrix exponential \(e^{\Delta t L}\), while the nonlinear term is handled by the ETD update. The resulting spectral components $\hat{\textbf{u}}_{n+1}$ are then recombined to recover the solution in the spatial domain, yielding $\mathbf{u}_{n+1}$.}
    \label{fig:diagrama}
\end{figure}

We introduce \textbf{\edpmethod{}}, an extension of the Neuro-Spectral Architecture (NeuSA) \cite{Bizzi2025NeuroSpectral} that integrates 4th-order Exponential Time Differencing into its temporal evolution. As in NeuSA, the nonlinear dynamics are learned, while known linear structure is treated explicitly, both in spectral space. Fig.~\ref{fig:diagrama} gives an overview of \edpmethod{}.

Recent work has explored incorporating first-order exponential integrators into the Neural ODE framework for simple stiff systems~\cite{fronk2025training, loya2025}, with extensions based on Taylor approximations~\cite{fronk2025taylor}.

In contrast to these approaches, 
our work integrates high-order ETD schemes into the Neural ODE and uses a spectral discretization to perform training on PDEs using a physical loss, yielding a unified formulation that substantially improves stability and computational efficiency in stiff PDE settings.

Let us consider an evolutionary PDE in the form \eqref{EDP}, where the differential operator is written as
\begin{equation}\label{eq:pde_split}
 \mathcal{F}( u ) = \mathcal{L}u  + \mathcal{N}( u )\,,
\end{equation}
where $\mathcal{L}$ is a linear partial differential operator and $\mathcal{N}$ denotes the remaining nonlinear term. Projecting $u$ onto an $n$-element spectral basis, we obtain an ODE system for the vector of coefficients 
$\hat{\mathbf{u}}$,
\begin{equation} \label{eq:discretized_PDE}
    \frac{d \hat{\mathbf{u}}}{dt} = \mathbf{L} \hat{\mathbf{u}} + \mathbf{N}(\hat{\mathbf{u}} )\,,
\end{equation}
where $\mathbf{L}:\mathbb{R}^{n}\mapsto \mathbb{R}^{n}$ is a linear diagonal vector field corresponding to the  linear and constant-coefficient part of $\mathcal{L}$ and $\mathbf{N}:\mathbb{R}^{n}\mapsto \mathbb{R}^{n}$ is a possibly nonlinear function, which may account for the variable-coefficient terms of a linear PDE \cite{Bizzi2025NeuroSpectral} together with the nonlinear terms in Eq.~\eqref{eq:pde_split}. In fact, this explicit separation is in many cases a convention which can be evaluated through analysis of the known differential equation with the objective of inserting most of the linear stiffness to the linear operator (see Appendix \ref{subsec:linsep}). 

In \edpmethod{}, the linear term $\mathbf{L} \hat{\mathbf{u}}$ is integrated exactly through ETDRK4~\citep{cox2002exponential}, as described in Appendix A.2, while the nonlinear term $\mathbf{N}(\hat{\mathbf{u}})$ is modeled by a neural network $\mathbf{N}_\theta(\hat{\mathbf{u}})$. 
We study the choice of a high-order integrator in Appendix C.1. Since $\mathbf{L}$ is diagonal, the matrix exponentials required for ETD are computationally efficient. The network is trained by minimizing a loss that can be defined in several ways, depending on the available information and the problem setting. The physics-based loss can be computed either in physical space or in Fourier space. In the latter case it reduces to a comparison of the nonlinear fields, i.e.,
\begin{equation}
\mathcal{L}_{NL}(\theta) = \sum_{n=1}^N \|\mathbf{N}_\theta(\hat{\mathbf{u}}_\theta(t_n)) - \mathbf{N}(\hat{\mathbf{u}}_\theta(t_n))\|^2,
\end{equation}
where $\hat{\mathbf{u}}_\theta$ is obtained by integrating the full vector field $\mathbf{F}_\theta = \mathbf{L} + \mathbf{N}_\theta$ using exponential time-differencing. In data-driven training, or when more complex boundary conditions are involved, the loss must instead be evaluated in physical space.

Such an approach produces better results under stiffness than the original NeuSA, since the linear term $\mathbf{L}$, which is often the source of stiffness in PDEs, is handled analytically, with only the nonlinear part being integrated with a Runge-Kutta-like solver (details in Appendix \ref{sec:ap_ETD}).

This extends the Neuro-Spectral framework to a substantially broader class of PDEs for which the original approach, using a Runge-Kutta solver, would require a prohibitive number of time-integration steps. At the same time, \edpmethod{} retains NeuSA's causality by construction and its mitigation of spectral bias, making it a competitive alternative to PINNs and related models.

In the following section, we evaluate \edpmethod{} on a set of PDE benchmarks that exhibit the stiffness and stiff-like behavior discussed above.

\section{Experimental Results}

Our experiments involve training \edpmethod{} for several forward initial/boundary value problems, and some applications in inverse problems.
Our baseline models are \textit{vanilla} PINNs \cite{raissi2019physics}, QRes \cite{bu2021qres} and FLS \cite{wong2024fls}. From now on, we will refer to the last three models as \textit{baseline models}. The adopted configurations of the baseline models are the same as in \cite{wu2024ropinn} and \cite{zhao2023pinnsformer}.
All of these experiments are stiff or otherwise stability-limited: at the step sizes used by \edpmethod{}, an explicit Runge--Kutta solver diverges numerically. NeuSA could in principle be trained with a much smaller step size; however, this would proportionally increase training cost and hinder convergence.

We evaluate the predicted solutions by computing the \textit{relative root mean squared error} (or \textit{the relative $L_2$ error}, defined in Appendix \ref{sec:ap_exp_setup}) with respect to a high-resolution reference solution obtained using an RK4 solver. 
The proposed and baseline methods are trained until the loss stabilizes, which defines the effective training time (TT). Each experiment is repeated over 10 different random seeds to control for outliers.

All experiments were executed on identical Ubuntu machines, each containing an Nvidia RTX 5090 GPU with 32GB VRAM, and an Intel i9-13900K processor with 128GB RAM. The quantitative results are displayed in Table \ref{table:results}.

\subsection{Forward problems}

Strong performance on forward problems indicates a model’s ability to accurately solve PDEs. It also provides confidence that the model can reliably interpolate sparse observations in a physically consistent manner, particularly in sparse data regimes where the physics-based loss dominates. To ensure the robustness of our model, we evaluate it on different forward problems.

\begin{table*}[h!]
\centering
\caption{\textbf{Quantitative evaluation on forward PDE benchmarks}. We report the mean relative root mean squared error (rRMSE) and training time (TT, in seconds) over 10 independent runs with different random seeds. \edpmethod{} consistently achieves lower error while requiring less training time than competing methods.
The standard deviations of these quantities are also shown in Appendix~\ref{sec:ap_exp_setup}.}
\setlength{\tabcolsep}{6pt}
\begin{tabular}{@{}lcc|cc|cc|cc|cc@{}}
\toprule
\multirow{2}{*}{\textbf{Model}} & \multicolumn{2}{c|}{\textbf{Heat 1d}} & \multicolumn{2}{c|}{\textbf{Burgers 1d}} & \multicolumn{2}{c|}{\textbf{Wave 1d}} & \multicolumn{2}{c|}{\textbf{KdV 1d}} & \multicolumn{2}{c}{\textbf{Heat 2d}} \\ \cmidrule(l){2-11} 
                                & \textbf{rRMSE}     & \textbf{TT}      & \textbf{rRMSE}      & \textbf{TT}        & \textbf{rRMSE}    & \textbf{TT}       & \textbf{rRMSE}    & \textbf{TT}      & \textbf{rRMSE}    & \textbf{TT}      \\ \midrule
PINN                            & 0.0346            & 157.94           & 0.0475             & 390.11             & 0.7789          & 429.63                           & 0.1411           & 1147.38          & 0.0810            & 297.14           \\
QRes                            & 0.0174            & 211.68           & 0.0228             & 544.32             & 0.7559          & 578.93                           & 0.0344           & 1562.83          & 0.0217            & 534.00           \\
FLS                             & 0.0485            & 163.58           & 0.0420             & 399.15             & 0.7312          & 441.11                           & 0.1496           & 1176.14          & 0.0674            & 304.23           \\
\textbf{\pdemethod{}}                       & \textbf{0.0030}   & \textbf{77.74}   & \textbf{0.0011}    & \textbf{174.28}    & \textbf{0.0031} & \textbf{135.50}                  & \textbf{0.0035}  & \textbf{944.86}  & \textbf{0.0010}   & \textbf{229.18}  \\ \bottomrule
\end{tabular}
\label{table:results}
\end{table*}

\subsubsection{1D Heat equation}

The heat equation is a prototypical stiff parabolic PDE, characterized by strong dissipation and widely separated temporal scales. Due to the quadratic growth of the eigenvalues in the spectral domain, explicit time integration schemes are severely constrained by stability requirements.

In our experiments, we consider the equation
\begin{align} \label{eq:heat1d}
\frac{\partial u}{\partial t} = \kappa \frac{\partial^2 u}{\partial x^2}\,,
\end{align}
for $(t,x) \in [0,2]\times[-2,2]$, with Gaussian initial condition and homogeneous Dirichlet boundary conditions. The thermal diffusivity profile $\kappa=\kappa(x)$ consists of two sharp sigmoid transitions located at $x=-0.5$ and $x=0.5$, creating three regions of increasing conductivity representing a sandwich structure of heat isolators and conductors (illustrated in Appendix~\ref{sec:ap_exp_setup}). The source is centered at $x=0.4$, near the right-most interface, in order to intentionally introduce asymmetry in the solution.

\begin{figure}[!htb]
\centering
\includegraphics[width=0.9\linewidth]{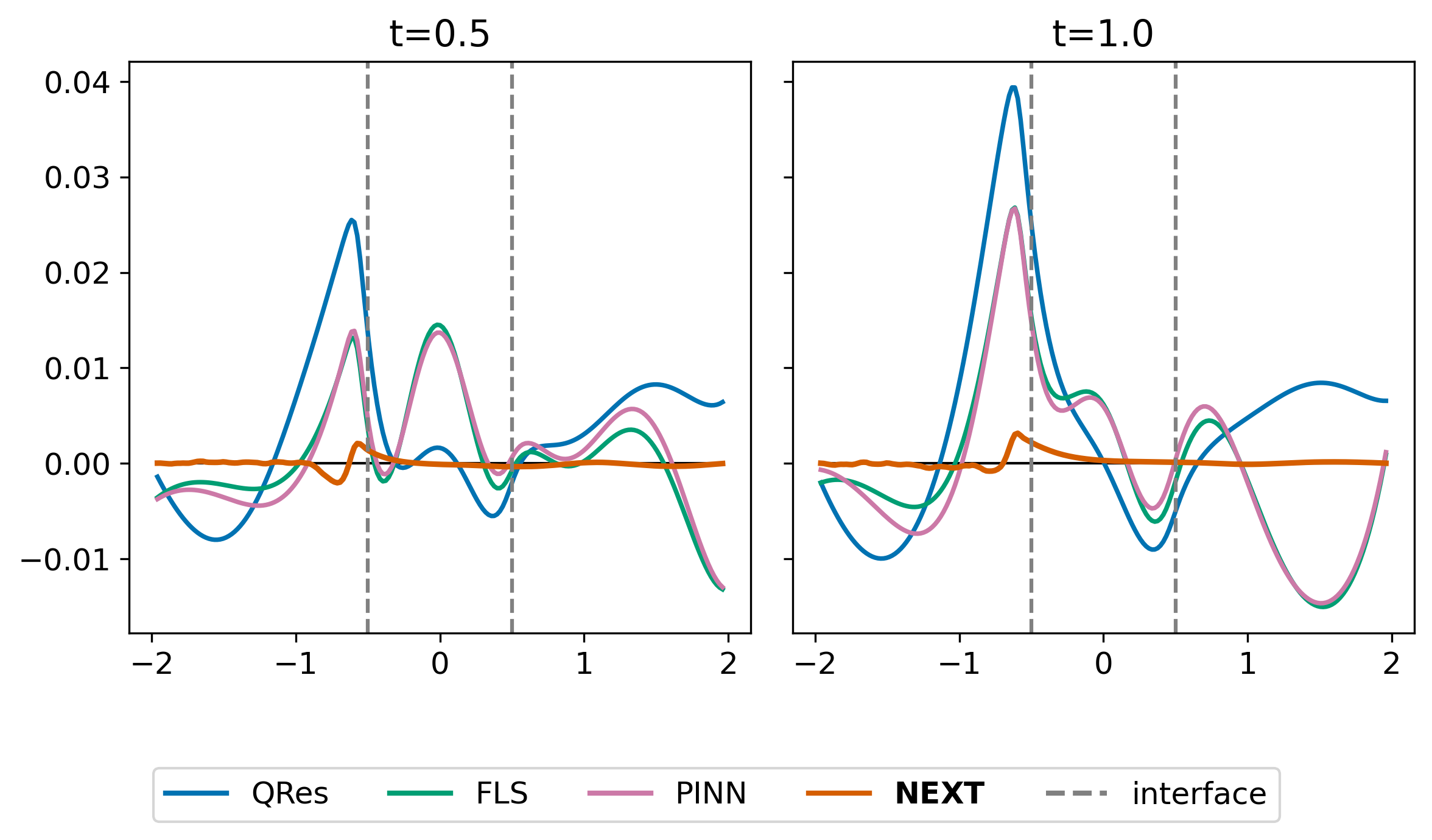}
\caption{Error comparison of the predicted solutions for the one-dimensional heat equation with heterogeneous conductivity.}
\label{fig:heat1d}
\end{figure}

As reported in Table \ref{table:results}, \edpmethod{} consistently outperforms the baseline models both in solution accuracy and training time. Fig.~\ref{fig:heat1d} shows that our model deals much better with changes in the thermal diffusivity profile.

\subsubsection{Viscous Burgers' equation}

Burgers' equation is a convection--diffusion equation arising in fluid dynamics, where the nonlinear convective term models advective transport, while the viscous term accounts for the diffusion. We consider the equation
\begin{equation}
\label{eq:burgers}
    \frac{\partial u}{\partial t} = - u\frac{\partial u}{\partial x} + \nu\frac{\partial^2 u}{\partial x^2}\,,
\end{equation}
where $(t,x) \in [0,1]\times[-1,1]$, $\nu=0.1$, with initial condition $u(0,x) = -\sin(\pi x)$, and periodic boundary condition. Although the viscous Burgers equation is not as strongly stiff as the heat equation, for $\nu = 0.1$ and sufficiently fine spatial discretizations it exhibits stiff behavior that severely restricts explicit time stepping schemes such as RK4.

Our model performs even better here than on the one-dimensional heat equation: \edpmethod{} outperforms the strongest baseline in Table~\ref{table:results} by more than an order of magnitude, at less than half its training time.

\subsubsection{Wave equation}\label{sub:wave}

Due to its causal nature, the wave equation represents a particularly difficult problem for vanilla PINNs without specialized training strategies \cite{zhongkai2024pinnacle, ding2024papermarcinho, mishra2025wave,marques2025}. 
Our one-dimensional wave equation is given by
\begin{equation}\label{we1d}
    \frac{\partial^2 u}{\partial t^2}
    = c^2 \frac{\partial^2 u}{\partial x^2}\,,
\end{equation}
where homogeneous Dirichlet boundary conditions and a Gaussian initial condition are imposed, and $c=c(x)$ is a heterogeneous velocity field, described in Appendix \ref{sec:ap_exp_setup}. To fit the first-order form of eq.~\eqref{eq:time_PDE}, we rewrite eq.~\eqref{we1d} as a system in $(u, u_t)$; see Appendix~\ref{sec:ap_exp_setup} for details.

Runge-Kutta solvers are typically well-suited for wave propagation problems; however, a balance between time and space discretization is required, as expressed by the Courant-Friedrichs-Lewy condition \cite{gottlieb1991cfl}. Thus, for a fixed space discretization, NeuSA with RK4 diverges unless a sufficiently small time step is used, whereas \edpmethod{} can take larger time steps, solving the 1D wave equation more efficiently. Furthermore, \edpmethod{} strongly outperforms all baselines, both in terms of accuracy and training time.

\begin{figure}[htb]
    \centering
    \includegraphics[width=0.9\linewidth]{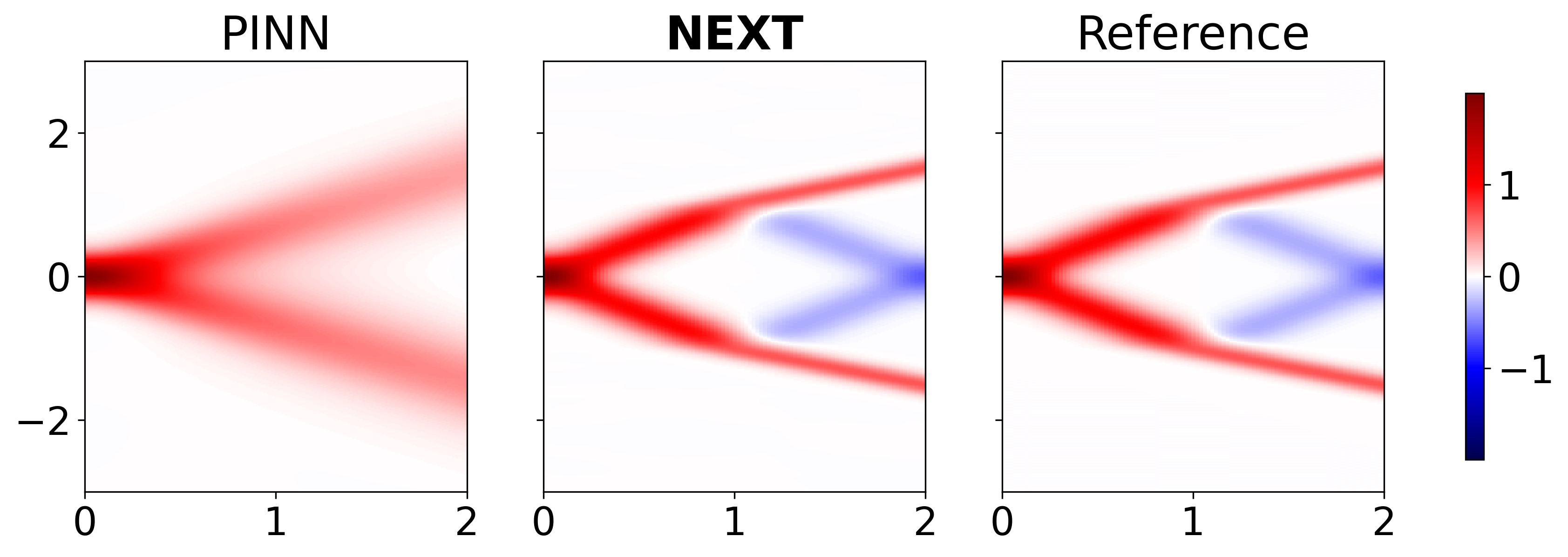}
    \caption{Solutions obtained by PINN and \pdemethod{} compared to the reference solution of the wave equation. Our method stays very close to the reference, while PINN fails to learn the reflected wave.}
    \label{fig:wave_solution}
\end{figure}

Fig.~\ref{fig:wave_solution} shows the solution obtained by \edpmethod{} compared to PINNs, which fail to capture the reflected wave entirely. 

\subsubsection{Korteweg--De Vries equation}

The considered KdV equation is similar to the formulation in \cite{zabusky1965interaction}, originally introduced in the seminal numerical study of soliton dynamics. For small dispersion parameters, the equation forms a nonlinear dispersive shock regime, which exhibits highly oscillatory dynamics and soliton formation, posing a challenge for PINNs. We tested the equation
\begin{equation}
    \label{eq:kdv}
    \frac{\partial u}{\partial t} =  \alpha u\frac{\partial u}{\partial x}+\delta\frac{\partial^3 u}{\partial x^3}\,,
\end{equation}
for $(t,x) \in [0,1]\times[0,2]$, $\alpha=-0.5$, $\delta=-0.022^2$, with initial condition $u(0,x) = \cos(\pi x)$ and periodic boundary conditions. 

\begin{figure}[h]
\centering
\includegraphics[scale=0.40]{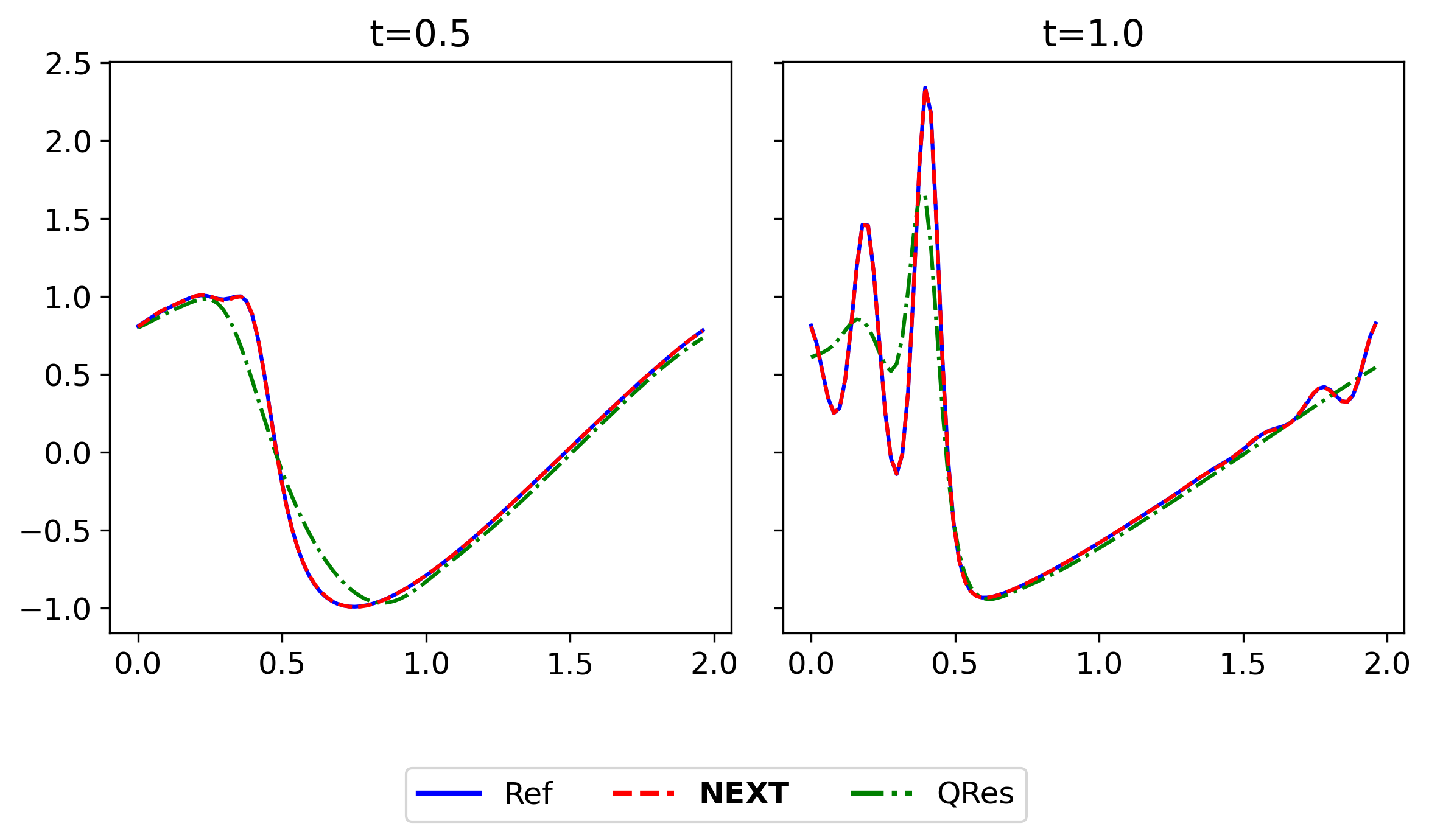}
\caption{Time snapshots of the \pdemethod{} and QRes solutions for the KdV equation in the dispersive shock regime.}
\label{fig:kdv_plot}
\end{figure}

\begin{figure*}[t]
  \centering
  \includegraphics[scale=0.35]{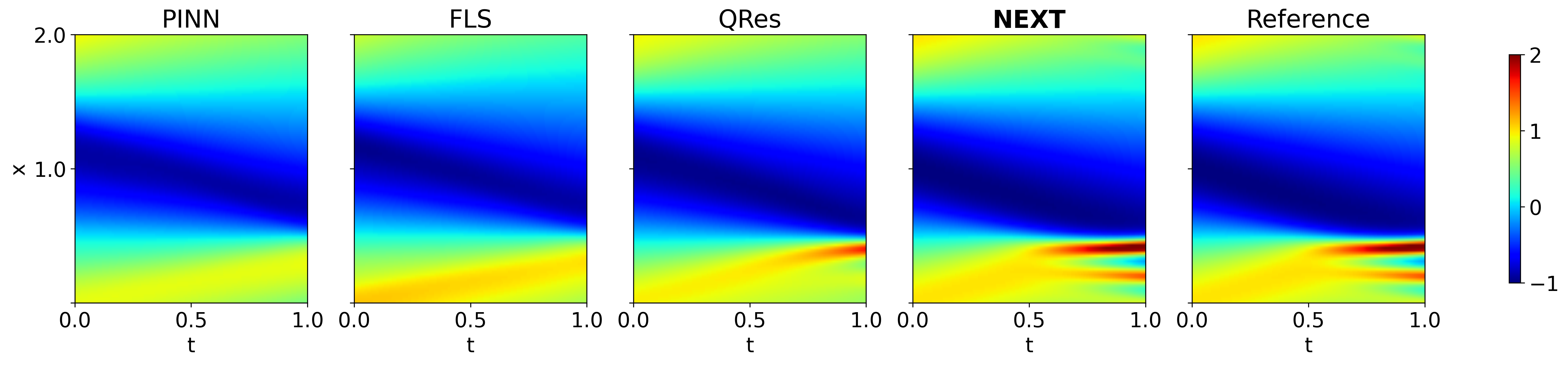}
\caption{Predicted solutions for the Korteweg--De Vries equation, illustrating that our model more accurately captures the nonlinear dispersive shock.}
  \label{fig:kdv}
\end{figure*}

In this problem, \edpmethod{} outperforms the baseline models by roughly an order of magnitude. As shown in Fig.~\ref{fig:kdv_plot} and Fig.~\ref{fig:kdv}, our model exhibits much closer agreement with the reference solution. Due to spectral bias, the baseline models were not able to capture the high-frequency components of the solution that become significant for $t > 0.5$, when a dispersive shock starts to develop.

\subsubsection{2D Heat equation}

We study a two-dimensional heat equation with spatially varying conductivity, which introduces heterogeneous diffusion effects absent in the one-dimensional setting.
Consider the equation
\begin{equation}
    \frac{\partial u }{\partial t}= \kappa \Delta u\,,
\end{equation}
where $\Delta$ denotes the standard Laplacian, for $(t,x,y) \in [0,2]\times[-2,2]\times[-2,2]$. The conductivity profile $\kappa = \kappa(x,y)$ represents a square-shaped region of high conductivity embedded in a low-conductivity medium. The initial condition concentrates the heat in a centrally located square-shaped region, and homogeneous Dirichlet boundary conditions are imposed.

From Table~\ref{table:results}, we observe that \edpmethod{} outperforms the baseline models by over an order of magnitude in accuracy while maintaining a lower training time. Unlike the baselines, \edpmethod{} captures the effect of the thermal insulator surrounding the high-conductivity region, as shown in Fig.~\ref{fig:teaser}.

\subsection{Inverse Problems}

PINNs are an appealing technique  to inverse problems, i.e., problems where one aims to recover unknown parameters or boundary conditions from sparse observations. However, in some problems the architecture's spectral bias and lack of causality degrade the solution away from the data. NeuSA partially addresses these issues, however, it fails in stiff regimes, where \pdemethod{} becomes a more suitable architecture.

\subsubsection{Burgers parameter identification}

As a test of \pdemethod{}'s capacity to identify PDE parameters from sparse solution data, we consider the inverse problem presented by \citet{raissi2019physics} for the Burgers equation, where we need to learn the unknown parameters $\lambda_1$ and $\lambda_2$ in a modification of eq. \eqref{eq:burgers}:
\begin{equation}
    \frac{\partial u}{\partial t} = - \lambda_1 u\frac{\partial u}{\partial x} + \lambda_2\frac{\partial^2 u}{\partial x^2}\, ,
\end{equation}
while having access to just two snapshots of the solution $u$, at $t=0$ and $t=1$.
We consider the same physical conditions as in the previous Burgers experiment, where $\lambda_1^{true} = 1$, $\lambda_2^{true} = \nu = 0.1$, and the solution domain is $(t,x) \in [0,1]\times [-1,1]$ with periodic boundary conditions.

As the linear field $\mathcal{L}u = \lambda_2 \partial_{xx} u$ now depends on a $\lambda_2$ that changes throughout training, the matrix exponentials required for ETD must now be computed at every training step. However, since $\mathbf{L}$ is diagonal, this incurs only a small computational cost.

Since the starting values for the learned PDE parameters can significantly affect the final results, we repeat the training experiment 10 times with different random seeds, randomly sampling $(\lambda_1^0, \lambda_2^0)$ uniformly on $[0.1, 2.0]\times[0.01, 0.20]$. The averaged results are reported in Table \ref{tab:inverse_results}. We do not report any results for NeuSA with the default Runge-Kutta solver because, for this stiff PDE, its RK4 integration diverges already at initialization.

Due to lack of causality, the baseline models have a much harder time than \pdemethod{} reconstructing the solution from two snapshots that are far apart in time, and consequently finding the true values of the parameters $\lambda_1$ and $\lambda_2$.

\begin{table}[ht]
\centering
{
\setlength{\tabcolsep}{2pt}
\caption{\textbf{Burgers parameter identification.} Mean and standard deviation of the learned parameters $\lambda_1,\, \lambda_2$ over 10 independent experiments, as well as the mean values of the rRMSE for the solution $u$ and the training time.}
    \label{tab:inverse_results}

\begin{tabular}{@{}lcccc@{}}
\toprule
\textbf{Model} & $\mathbf{\lambda_1}$       & $\mathbf{\lambda_2}$        & \textbf{rRMSE} & \textbf{TT} \\ \midrule
PINN                               & $0.506 \pm 0.201$ & $0.108 \pm 0.003$ & 0.099          & 405         \\
QRes                               & $0.805 \pm 0.103$ & $0.104 \pm 0.002$ & 0.040          & 548         \\
FLS                                & $0.509 \pm 0.203$ & $0.108 \pm 0.004$ & 0.099          & 407         \\
\textbf{\pdemethod{}}                               & $\mathbf{0.955 \pm 0.041}$ & ${\mathbf{0.101} \pm \mathbf{0.001}}$ & $\mathbf{0.009}$          & $\mathbf{209}$         \\ \midrule
True Value                       & 1.000                 & 0.100               & -              & -           \\ \bottomrule
\end{tabular}

}
\end{table}

\subsubsection{Boundary heat flux inversion}

In extreme environments where directly measuring  external thermal conditions may not be feasible, the inverse heat conduction problems (IHCPs) have been widely used to determine  thermal boundary conditions ~\cite{wen2023real,lee2025heat}. As demonstrated  by \cite{shang2025simultaneous}, PINNs can be a useful framework for solving such problems. To demonstrate the flexibility of \edpmethod{} in practical problems, we solve an inverse heat flux problem based on an industrial application~\cite{shang2025simultaneous}.

We consider a one-dimensional domain of length $L_x = 0.02 \mbox{ m}$ subject to a transient 
heat flux at the left boundary by $t_f = 1000 \mbox{ s}$, in addition to convection and radiation heat 
transfer at the right boundary.  The governing equation 
for this problem is:
\begin{equation}
    \begin{cases}
        T_{t}(x,t) = \dfrac{k}{\rho c_p} T_{xx}(x,t) \\[10pt]
        T(x,0) = T_0 \\[6pt]
        -k T_{x}(x,t)\big|_{x=0} = q_{\text{in}}(t) 
        - \sigma \varepsilon_{w1}\left(T_{w1}^4 - T_a^4\right) \\[6pt]
        -k T_{x}(x,t)\big|_{x=L_x} = h(T_{w2} - T_a) 
        + \sigma \varepsilon_{w2}\left(T_{w2}^4 - T_a^4\right)
    \end{cases}
    \label{eq:governing}
\end{equation}
where $\rho$ is the material density; $c_p$ is the  specific heat capacity; $k$ is the thermal 
conductivity given by a linear function of the temperature $k(T) = k_0 - k_1T$; $T(x,t)$ is the 
temperature field; $T_0$ is the uniform initial temperature; $T_t$, $T_x$ and
$T_{xx}$ denote the first-order time derivative, the first-order and the second-order spatial 
derivative of the temperature, respectively; $q_{\text{in}}(t)$ is the 
unknown boundary heat flux at the left end; $h$ is the convection heat 
transfer coefficient and $T_a$ is the ambient temperature at the right end; 
$\sigma$ is the Stefan--Boltzmann constant; $\varepsilon$ is the radiation emissivities at the left and right 
boundaries, respectively; and $T_{w1}$, $T_{w2}$ are the surface temperatures 
at the left and right ends. The values for the thermophysical parameters are listed in Appendix \ref{sec:ap_exp_setup}.

\begin{figure}[h]
\centering
\includegraphics[width=0.9\linewidth]{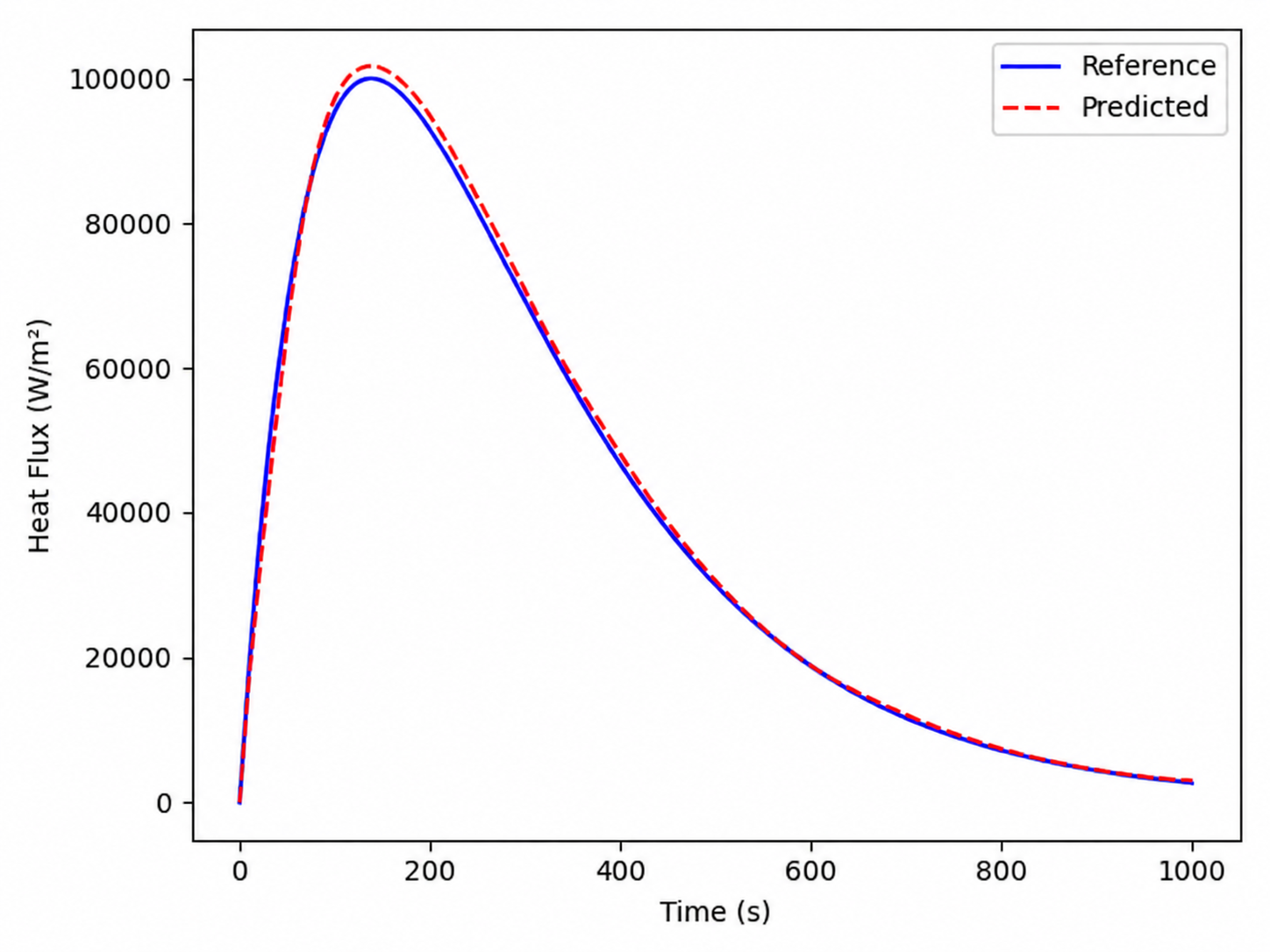}
\caption{Predicted heat flux on the inverse heat problem using \pdemethod{}.}
\label{fig:inverse_heat}
\end{figure}

Transient temperature data at the locations $x = L_x/2$ and $x = L_x$, 
where the two temperature sensors are placed, are extracted and denoted as 
$T_1$ and $T_2$, respectively. These measurements serve as inputs to the 
inverse problem, whose goal is to recover the unknown boundary heat flux 
$q_{\text{in}}(t)$. The heat flux profile considered in this problem is:
\begin{equation}
    q_{\text{in}}(t) = q_{0}\left(1 - e^{-\omega_1 t}\right)e^{-\omega_1 t}
    \label{eq:q1}
\end{equation}
where $q_{0} = 4 \times 10^5~\text{W/m}^2$ is the heat flux amplitude and 
$\omega_1 = 0.005~\text{s}^{-1}$ is the decay rate parameter.
The inverse heat conduction problem highlights the strength of \edpmethod{} in reconstructing boundary heat flux from spatially sparse measurements. Even with limited sensor data, \edpmethod{} accurately recovers the temperature field throughout the entire domain, attaining a relative RMSE of 0.035 for the heat flux, 0.0021 for the temperature, and 0.0015 for the sensor measurements, with a training time of 389 seconds. As shown in Fig.~\ref{fig:inverse_heat}, the predicted heat flux closely follows the reference solution, confirming that accurate boundary information can be inferred from interior observations alone. More broadly, these results illustrate the promise of \edpmethod{} as a prominent tool in scientific machine learning, especially for ill-posed inverse problems where direct boundary measurements are difficult or impossible to obtain.

\subsection{Results and Discussion}

Due to stiffness, standard NeuSA models \cite{Bizzi2025NeuroSpectral} are not suited for the heat, Burgers and Korteweg--De Vries equations being dealt with here, or for the presented wave problem. Thus, we conclude that adopting exponential time-differencing methods enables us to solve a much larger class of problems. 

As seen in Table~\ref{table:results}, \edpmethod{} outperforms the strongest baseline on every equation tested, by factors ranging from roughly $6\times$ (1D heat) to over $200\times$ (1D wave), while consistently requiring shorter training times. For the heat and the Burgers equations, in which PINNs traditionally achieve strong results, \edpmethod{} converges to a much more accurate solution in the same training time. For the wave and KdV PDEs, the MLPs miss important details in the solution regardless of training time, while \edpmethod{} follows the solution very closely, as seen in Figures \ref{fig:wave_solution} and \ref{fig:kdv}.

Both inverse experiments demonstrate \edpmethod{}'s practical utility for parameter identification. In the Burgers problem, the method accurately recovers the unknown PDE parameters, as seen in Table~\ref{tab:inverse_results}. In the inverse heat conduction problem, it successfully reconstructs an unknown transient boundary heat flux from just two temperature sensors, as in Fig. ~\ref{fig:inverse_heat}. Together, these results show that \edpmethod{} handles both unknown coefficients and unknown boundary conditions reliably and from very few measurements, making it an effective and flexible tool for real-world inverse problems.

\section{Conclusion}

NeuSA is a promising method for learning an implicit neural representation of PDE solutions through physics-informed training, while avoiding spectral bias and preserving causality, which are difficult for vanilla PINNs. However, in its original formulation, it had limitations when applied to stiff PDEs, for which the Runge-Kutta numerical solver made it unstable. Our contribution of combining NeuSA with a fourth-order exponential solver, resulting in \edpmethod{}, allows us to generalize it to a much broader class of PDEs.

In our approach, the possibly stiff linear terms of the PDE are integrated exactly through ETD, while the nonlinear part is represented by a neural network, thus allowing us to benefit from the PDE structure and obtain fast and accurate solutions as compared to the baselines.

This work opens many directions of future research and applications. First, \edpmethod{} can be applied to a wider set of equations. Although the formulation is presented of autonomous PDEs, the heat inverse problem shows that \edpmethod{} is actually capable of dealing with non-autonomous PDEs as well. In future work we also plan to address more challenging problems, including the Navier-Stokes equations \cite{batchelor2005book}, elasticity equations and the models of multi-phase flows that require dynamical updates of the vector fields via the solution of an elliptic equation. It could be also interesting to extend the proposed technique to the case of pseudo-differential equations \cite{acosta2024pseudodiff,CSTOLK2004pseudodiff} that are important, e.g., for simulating tsunami waves. 

Further improvements of the \edpmethod{} approach can likely be achieved by using other existing families of high-order exponential integrators \cite{krogstad2005generalized,asante2025fourth}, many of which have important advantages for specific classes of equations. Another important issue on the side of computational mathematics to be resolved within the \edpmethod{} framework is artificial domain truncation by perfectly matched layers or transparent boundary conditions \cite{antoine2008review}.

\newpage
\bibliography{refs}

\begin{thebibliography}{56}
\providecommand{\natexlab}[1]{#1}
\providecommand{\url}[1]{\texttt{#1}}
\expandafter\ifx\csname urlstyle\endcsname\relax
  \providecommand{\doi}[1]{doi: #1}\else
  \providecommand{\doi}{doi: \begingroup \urlstyle{rm}\Url}\fi

\bibitem[Acosta et~al.(2024)Acosta, Chan, Johnson, and
  Palacios]{acosta2024pseudodiff}
Sebastian Acosta, Jesse Chan, Raven Johnson, and Benjamin Palacios.
\newblock Pseudodifferential models for ultrasound waves with fractional
  attenuation.
\newblock \emph{SIAM Journal on Applied Mathematics}, 84\penalty0 (4), 2024.
\newblock \doi{10.1137/24M1634011}.

\bibitem[Antoine et~al.(2008)Antoine, Arnold, Besse, Ehrhardt, and
  Sch{\"a}dle]{antoine2008review}
Xavier Antoine, Anton Arnold, Christophe Besse, Matthias Ehrhardt, and Achim
  Sch{\"a}dle.
\newblock A review of transparent and artificial boundary conditions techniques
  for linear and nonlinear schr{\"o}dinger equations.
\newblock \emph{Communications in computational physics}, 4\penalty0
  (4):\penalty0 729--796, 2008.

\bibitem[Asante-Asamani et~al.(2025)Asante-Asamani, Kleefeld, and
  Wade]{asante2025fourth}
EO~Asante-Asamani, Andreas Kleefeld, and Bruce~A Wade.
\newblock A fourth-order exponential time differencing scheme with dimensional
  splitting for non-linear reaction--diffusion systems.
\newblock \emph{Journal of Computational and Applied Mathematics},
  465:\penalty0 116568, 2025.

\bibitem[Balakrishnan et~al.(2019)Balakrishnan, Zhao, Sabuncu, Guttag, and
  Dalca]{balakrishnan2019voxelmorph}
Guha Balakrishnan, Amy Zhao, Mert~R Sabuncu, John Guttag, and Adrian~V Dalca.
\newblock Voxelmorph: a learning framework for deformable medical image
  registration.
\newblock \emph{IEEE transactions on medical imaging}, 2019.

\bibitem[Balestro et~al.(2026)Balestro, Marques, Mendon{\c{c}}a, Moreira,
  de~Oliveira, Ganacim, Novello, Petrov, Yukimura, and
  Nissenbaum]{balestro2026neurospectral}
Vitor Balestro, M{\'a}rcio Marques, Leonardo Mendon{\c{c}}a, Leonardo~M.
  Moreira, Christian~J{\'u}nior de~Oliveira, Francisco Ganacim, Tiago Novello,
  Pavel Petrov, Daniel Yukimura, and Lucas Nissenbaum.
\newblock Neuro-spectral architectures with time-domain decomposition.
\newblock In \emph{AI{\&}PDE: ICLR 2026 Workshop on AI and Partial Differential
  Equations}, 2026.
\newblock URL \url{https://openreview.net/forum?id=sn5fl11bA0}.

\bibitem[Batchelor(2005)]{batchelor2005book}
G.~K Batchelor.
\newblock \emph{An introduction to fluid dynamics}.
\newblock Cambridge mathematical library. Cambridge University Press, reprint
  edition, 2005.

\bibitem[Bizzi et~al.(2025{\natexlab{a}})Bizzi, Grynberg, Matias, Perazzo,
  Lima, Velho, Gonçalves, Pereira, Schardong, and Novello]{novello2025}
Arthur Bizzi, Matias Grynberg, Vitor Matias, Daniel Perazzo, João~Paulo Lima,
  Luiz Velho, Nuno Gonçalves, João Pereira, Guilherme Schardong, and Tiago
  Novello.
\newblock Flowing: Implicit neural flows for structure-preserving morphing.
\newblock In \emph{Conference on Neural Information Processing Systems
  (NeurIPS)}, volume~38, 2025{\natexlab{a}}.

\bibitem[Bizzi et~al.(2025{\natexlab{b}})Bizzi, Moreira, Marques, Mendonça,
  de~Oliveira, Balestro, Fernandez, Yukimura, Petrov, Pereira, Novello, and
  Nissenbaum]{Bizzi2025NeuroSpectral}
Arthur Bizzi, Leonardo~M. Moreira, Márcio Marques, Leonardo Mendonça,
  Christian~Júnior de~Oliveira, Vitor Balestro, Lucas dos~Santos Fernandez,
  Daniel Yukimura, Pavel Petrov, João~M. Pereira, Tiago Novello, and Lucas
  Nissenbaum.
\newblock Neuro-spectral architectures for causal physics-informed networks.
\newblock In \emph{Conference on Neural Information Processing Systems
  (NeurIPS)}, 2025{\natexlab{b}}.

\bibitem[Bu and Karpatne(2021)]{bu2021qres}
Jie Bu and Anuj Karpatne.
\newblock Quadratic residual networks: A new class of neural networks for
  solving forward and inverse problems in physics involving pdes.
\newblock In \emph{Proceedings of the 2021 SIAM International Conference on
  Data Mining (SDM)}, pages 675--683. SIAM, 2021.

\bibitem[Burden et~al.(2016)Burden, Faires, and
  Burden]{burden2016_numerical_analysis}
Richard~L. Burden, J.~Douglas Faires, and Annette~M. Burden.
\newblock \emph{Numerical Analysis}.
\newblock Cengage Learning, Boston, MA, 10th edition, 2016.
\newblock ISBN 978-1-305-25366-7.

\bibitem[Canuto et~al.(2007)Canuto, Hussaini, Quarteroni, and
  Zang]{canuto2007spectral}
Claudio Canuto, M.~Yousuff Hussaini, Alfio Quarteroni, and Thomas~A. Zang.
\newblock \emph{Spectral Methods: Evolution to Complex Geometries and
  Applications to Fluid Dynamics}.
\newblock Springer, 2007.
\newblock \doi{10.1007/978-3-540-30726-6}.

\bibitem[Chandra and Kapoor(2026)]{chandra2026oscillatory}
Abhishek Chandra and Taniya Kapoor.
\newblock Oscillatory state-space models as inductive biases for
  physics-informed neural pde solvers, 2026.
\newblock URL \url{https://arxiv.org/abs/2606.02623}.

\bibitem[Chen et~al.(2018)Chen, Rubanova, Bettencourt, and
  Duvenaud]{chen2019neural}
Ricky T.~Q. Chen, Yulia Rubanova, Jesse Bettencourt, and David Duvenaud.
\newblock Neural ordinary differential equations.
\newblock In \emph{Proceedings of the 32nd International Conference on Neural
  Information Processing Systems}, NIPS'18, page 6572–6583, Red Hook, NY,
  USA, 2018. Curran Associates Inc.

\bibitem[Cox and Matthews(2002)]{cox2002exponential}
Steven~M Cox and Paul~C Matthews.
\newblock Exponential time differencing for stiff systems.
\newblock \emph{Journal of Computational Physics}, 176\penalty0 (2):\penalty0
  430--455, 2002.

\bibitem[De~Ryck et~al.(2022)De~Ryck, Mishra, and Molinaro]{de2022weak}
Tim De~Ryck, Siddhartha Mishra, and Roberto Molinaro.
\newblock Weak physics informed neural networks for approximating entropy
  solutions of hyperbolic conservation laws.
\newblock In \emph{Seminar f{\"u}r Angewandte Mathematik, Eidgen{\"o}ssische
  Technische Hochschule, Z{\"u}rich, Switzerland, Rep}, volume~35, page 2022,
  2022.

\bibitem[Ding et~al.(2025)Ding, Chen, Miyake, and Li]{ding2024papermarcinho}
Yi~Ding, Su~Chen, Hiroe Miyake, and Xiaojun Li.
\newblock Physics-informed neural networks with fourier features for seismic
  wavefield simulation in time-domain nonsmooth complex media.
\newblock \emph{IEEE Transactions on Geoscience and Remote Sensing},
  63:\penalty0 1--13, 2025.
\newblock \doi{10.1109/TGRS.2025.3581638}.

\bibitem[Donnelly et~al.(2024)Donnelly, Daneshkhah, and
  Abolfathi]{donnelly2024hydro}
James Donnelly, Alireza Daneshkhah, and Soroush Abolfathi.
\newblock Physics-informed neural networks as surrogate models of hydrodynamic
  simulators.
\newblock \emph{Science of The Total Environment}, 912:\penalty0 168814, 2024.

\bibitem[Dupont et~al.(2019)Dupont, Doucet, and Teh]{dupont2019augmented}
Emilien Dupont, Arnaud Doucet, and Yee~Whye Teh.
\newblock Augmented neural odes.
\newblock \emph{Advances in neural information processing systems}, 32, 2019.

\bibitem[Fornberg and Driscoll(1999)]{fornberg1999nonlinearwave}
Bengt Fornberg and Tobin~A. Driscoll.
\newblock A fast spectral algorithm for nonlinear wave equations with linear
  dispersion.
\newblock \emph{Journal of Computational Physics}, 155\penalty0 (2):\penalty0
  456--467, 1999.
\newblock ISSN 0021-9991.
\newblock \doi{https://doi.org/10.1006/jcph.1999.6351}.
\newblock URL
  \url{https://www.sciencedirect.com/science/article/pii/S0021999199963519}.

\bibitem[Fronk and Petzold(2025{\natexlab{a}})]{fronk2025taylor}
Colby Fronk and Linda Petzold.
\newblock Training stiff neural ordinary differential equations with explicit
  rational taylor series methods.
\newblock \emph{Chaos: An Interdisciplinary Journal of Nonlinear Science},
  35\penalty0 (7), 2025{\natexlab{a}}.

\bibitem[Fronk and Petzold(2025{\natexlab{b}})]{fronk2025training}
Colby Fronk and Linda Petzold.
\newblock Training stiff neural ordinary differential equations with explicit
  exponential integration methods.
\newblock \emph{Chaos: An Interdisciplinary Journal of Nonlinear Science},
  35\penalty0 (3), 2025{\natexlab{b}}.

\bibitem[Gottlieb and Tadmor(1991)]{gottlieb1991cfl}
David Gottlieb and Eitan Tadmor.
\newblock The cfl condition for spectral approximations to hyperbolic
  initial-boundary value problems.
\newblock \emph{Mathematics of Computation}, 56\penalty0 (194):\penalty0
  565--588, 1991.

\bibitem[Hochbruck and Ostermann(2010)]{hochbruck2010exponential}
Marlis Hochbruck and Alexander Ostermann.
\newblock Exponential integrators.
\newblock \emph{Acta Numerica}, 19:\penalty0 209--286, 2010.

\bibitem[Hou et~al.(2024)Hou, Li, Singh, Sun, and Wei]{hou2024kinematicwave}
Qingzhi Hou, Yixin Li, Vijay~P. Singh, Zewei Sun, and Jianguo Wei.
\newblock Physics-informed neural network for solution of forward and inverse
  kinematic wave problems.
\newblock \emph{Journal of Hydrology}, 633:\penalty0 130934, 2024.
\newblock ISSN 0022-1694.
\newblock \doi{https://doi.org/10.1016/j.jhydrol.2024.130934}.
\newblock URL
  \url{https://www.sciencedirect.com/science/article/pii/S0022169424003287}.

\bibitem[Kassam and Trefethen(2005)]{kassam2005etd}
Aly-Khan Kassam and Lloyd~N. Trefethen.
\newblock Fourth-order time-stepping for stiff {PDE}s.
\newblock \emph{SIAM Journal on Scientific Computing}, 26\penalty0
  (4):\penalty0 1214--1233, 2005.
\newblock \doi{10.1137/S1064827502410633}.

\bibitem[Kidger(2022)]{kidger2022neural}
Patrick Kidger.
\newblock On neural differential equations, 2022.

\bibitem[Krogstad(2005)]{krogstad2005generalized}
Stein Krogstad.
\newblock Generalized integrating factor methods for stiff pdes.
\newblock \emph{Journal of Computational Physics}, 203\penalty0 (1):\penalty0
  72--88, 2005.

\bibitem[Lee et~al.(2025)Lee, Ahn, Park, Kim, Park, Park, and Kim]{lee2025heat}
Ji-won Lee, Chang-uk Ahn, Jongwoo Park, Hwi-su Kim, Chanhun Park, Dong~Il Park,
  and Jin-Gyun Kim.
\newblock Heat source and internal temperature estimation of an integrated
  modular motor drive in robotic application using inverse heat conduction
  problem.
\newblock \emph{Measurement}, 243:\penalty0 116297, 2025.

\bibitem[LeVeque(2007)]{leveque2007finite}
Randall~J. LeVeque.
\newblock \emph{Finite Difference Methods for Ordinary and Partial Differential
  Equations: Steady-State and Time-Dependent Problems}.
\newblock Society for Industrial and Applied Mathematics, 2007.
\newblock \doi{10.1137/1.9780898717839}.

\bibitem[Loya et~al.(2025)Loya, Serino, Burby, and Tang]{loya2025}
Allen~Alvarez Loya, Daniel~A. Serino, J.~W. Burby, and Qi~Tang.
\newblock Structure-preserving neural ordinary differential equations for stiff
  systems, 2025.

\bibitem[Marques et~al.(2025)Marques, Mendon{\c{c}}a, Bizzi, Moreira, Oliveira,
  Oliveira, Fernandez, Balestro, Pereira, Yukimura, Novello, Petrov, and
  Nissenbaum]{marques2025}
M{\'a}rcio Marques, Leonardo Mendon{\c{c}}a, Arthur Bizzi, Leonardo Moreira,
  Christian Oliveira, Deborah Oliveira, Lucas Fernandez, Vitor Balestro,
  Jo{\~a}o Pereira, Daniel Yukimura, Tiago Novello, Pavel Petrov, and Lucas
  Nissenbaum.
\newblock Stable adaptive training for physics-informed neural networks in
  acoustic wave propagation.
\newblock \emph{JASA Express Letters}, 5\penalty0 (11), 2025.

\bibitem[Mendon{\c{c}}a et~al.(2026)Mendon{\c{c}}a, Marques, Petrov, and
  Nissenbaum]{mendonca2026physicsinformed}
Leonardo Mendon{\c{c}}a, M{\'a}rcio Marques, Pavel Petrov, and Lucas
  Nissenbaum.
\newblock Physics-informed adaptive training for 3d acoustic wave propagation.
\newblock In \emph{AI{\&}PDE: ICLR 2026 Workshop on AI and Partial Differential
  Equations}, 2026.
\newblock URL \url{https://openreview.net/forum?id=DzFZMVs0B4}.

\bibitem[Mishra and Shekar(2025)]{mishra2025wave}
S~Mishra and B~Shekar.
\newblock Wave equation and neural solvers: What works and what needs to
  change?
\newblock In \emph{86th EAGE Annual Conference \& Exhibition}, volume 2025,
  pages 1--5. European Association of Geoscientists \& Engineers, 2025.

\bibitem[Molina~Catricheo et~al.(2024)Molina~Catricheo, Lambert, Salomon, and
  van~’t Wout]{molina2024modeling}
Constanza~A. Molina~Catricheo, Fabrice Lambert, Julien Salomon, and Elwin
  van~’t Wout.
\newblock Modeling global surface dust deposition using physics-informed neural
  networks.
\newblock \emph{Communications Earth \& Environment}, 5\penalty0 (1):\penalty0
  778, 2024.
\newblock \doi{10.1038/s43247-024-01942-2}.

\bibitem[Ortega(1990)]{ortega1990_numerical_analysis_second_course}
James~M. Ortega.
\newblock \emph{Numerical Analysis: A Second Course}, volume~3 of
  \emph{Classics in Applied Mathematics}.
\newblock Society for Industrial and Applied Mathematics, Philadelphia, 1990.
\newblock ISBN 978-0-89871-250-6.

\bibitem[Owoyele and Pal(2022)]{Owoyele2022}
Opeoluwa Owoyele and Pinaki Pal.
\newblock Chemnode: A neural ordinary differential equations framework for
  efficient chemical kinetic solvers.
\newblock \emph{Energy and AI}, 7:\penalty0 100118, 2022.
\newblock ISSN 2666-5468.

\bibitem[Patel et~al.(2022)Patel, Manickam, Trask, Lee, Tomas, and
  Cyr]{patel2022thermodynamically}
Ravi~G Patel, Indu Manickam, Mitchell~A. Trask, Nathaniel A.and~Wood, Myoungkyu
  Lee, Ignacio Tomas, and Eric~C Cyr.
\newblock Thermodynamically consistent physics-informed neural networks for
  hyperbolic systems.
\newblock \emph{Journal of Computational Physics}, 449:\penalty0 110754, 2022.

\bibitem[Raissi et~al.(2019)Raissi, Perdikaris, and
  Karniadakis]{raissi2019physics}
Maziar Raissi, Paris Perdikaris, and George~E Karniadakis.
\newblock Physics-informed neural networks: A deep learning framework for
  solving forward and inverse problems involving nonlinear partial differential
  equations.
\newblock \emph{Journal of Computational Physics}, 378:\penalty0 686--707,
  2019.

\bibitem[Shang et~al.(2025)Shang, Ban, and Liu]{shang2025simultaneous}
Yuanbin Shang, Huaiguo Ban, and Donghuan Liu.
\newblock Simultaneous identification of boundary heat flux and thermal
  conductivity in inverse heat conduction problems using physics-informed
  neural networks.
\newblock \emph{Thermal Science and Engineering Progress}, page 103905, 2025.

\bibitem[Stoer et~al.(1980)Stoer, Bulirsch, Bartels, Gautschi, and
  Witzgall]{stoer1980introduction}
Josef Stoer, Roland Bulirsch, R~Bartels, Walter Gautschi, and Christoph
  Witzgall.
\newblock \emph{Introduction to numerical analysis}, volume 1993.
\newblock Springer, 1980.

\bibitem[Stolk(2004)]{CSTOLK2004pseudodiff}
Christiaan Stolk.
\newblock A pseudodifferential equation with damping for one-way wave
  propagation in inhomogeneous acoustic media.
\newblock \emph{Wave Motion}, 40\penalty0 (2):\penalty0 111--121, 2004.
\newblock \doi{10.1016/j.wavemoti.2003.12.016}.

\bibitem[Sun et~al.(2022)Sun, Han, Kong, Tang, Yan, and Xie]{sun2022topology}
Shanlin Sun, Kun Han, Deying Kong, Hao Tang, Xiangyi Yan, and Xiaohui Xie.
\newblock Topology-preserving shape reconstruction and registration via neural
  diffeomorphic flow.
\newblock In \emph{CVPR}, 2022.

\bibitem[Sun et~al.(2024)Sun, Han, You, Tang, Kong, Naushad, Yan, Ma, Khosravi,
  Duncan, et~al.]{sun2024medical}
Shanlin Sun, Kun Han, Chenyu You, Hao Tang, Deying Kong, Junayed Naushad,
  Xiangyi Yan, Haoyu Ma, Pooya Khosravi, James~S Duncan, et~al.
\newblock Medical image registration via neural fields.
\newblock \emph{Medical Image Analysis}, 97:\penalty0 103249, 2024.

\bibitem[Trefethen(2000)]{trefethen2000spectral}
Lloyd~N. Trefethen.
\newblock \emph{Spectral Methods in MATLAB}.
\newblock Society for Industrial and Applied Mathematics, 2000.
\newblock \doi{10.1137/1.9780898719598}.

\bibitem[Wang et~al.(2022)Wang, Yu, and Perdikaris]{wang2022-pinnsfail}
Sifan Wang, Xinling Yu, and Paris Perdikaris.
\newblock When and why pinns fail to train: A neural tangent kernel
  perspective.
\newblock \emph{Journal of Computational Physics}, 449:\penalty0 110768, 2022.

\bibitem[Wang et~al.(2024)Wang, Sankaran, and Perdikaris]{wang2024causality}
Sifan Wang, Shyam Sankaran, and Paris Perdikaris.
\newblock Respecting causality for training physics-informed neural networks.
\newblock \emph{Computer Methods in Applied Mechanics and Engineering},
  421:\penalty0 116813, 2024.

\bibitem[Wen et~al.(2023)Wen, Ma, Zhou, and Sun]{wen2023real}
Shuang Wen, Yicheng Ma, Tian Zhou, and Zhiqiang Sun.
\newblock Real-time estimation of thermal boundary conditions and internal
  temperature fields for thermal protection system of aerospace vehicle via
  temperature sequence.
\newblock \emph{International Communications in Heat and Mass Transfer},
  142:\penalty0 106618, 2023.

\bibitem[Wong et~al.(2024)Wong, Ooi, Gupta, and Ong]{wong2024fls}
Jian~Cheng Wong, Chin~Chun Ooi, Abhishek Gupta, and Yew-Soon Ong.
\newblock Learning in sinusoidal spaces with physics-informed neural networks.
\newblock \emph{IEEE Transactions on Artificial Intelligence}, 5\penalty0
  (3):\penalty0 985--1000, 2024.
\newblock \doi{10.1109/TAI.2022.3192362}.

\bibitem[Wu et~al.(2023)Wu, Zhu, Tan, Kartha, and Lu]{wu2023-rad}
Chenxi Wu, Min Zhu, Qinyang Tan, Yadhu Kartha, and Lu~Lu.
\newblock A comprehensive study of non-adaptive and residual-based adaptive
  sampling for physics-informed neural networks.
\newblock \emph{Computer Methods in Applied Mechanics and Engineering},
  403:\penalty0 115671, 2023.

\bibitem[Wu et~al.(2024)Wu, Luo, Ma, Wang, and Long]{wu2024ropinn}
Haixu Wu, Huakun Luo, Yuezhou Ma, Jianmin Wang, and Mingsheng Long.
\newblock {R}o{PINN}: Region optimized physics-informed neural networks.
\newblock In A.~Globerson, L.~Mackey, D.~Belgrave, A.~Fan, U.~Paquet,
  J.~Tomczak, and C.~Zhang, editors, \emph{Advances in Neural Information
  Processing Systems}, volume~37, pages 110494--110532. Curran Associates,
  Inc., 2024.

\bibitem[Wu et~al.(2022)Wu, Jiahao, Wang, Yushkevich, Hsieh, and
  Gee]{wu2022nodeo}
Yifan Wu, Tom~Z Jiahao, Jiancong Wang, Paul~A Yushkevich, M~Ani Hsieh, and
  James~C Gee.
\newblock Nodeo: A neural ordinary differential equation based optimization
  framework for deformable image registration.
\newblock In \emph{CVPR}, 2022.

\bibitem[Xu et~al.(2019)Xu, Zhang, Luo, Xiao, and Ma]{xu2019frequencyprinciple}
Zhi-Qin~John Xu, Yaoyu Zhang, Tao Luo, Yanyang Xiao, and Zheng Ma.
\newblock Frequency principle: Fourier analysis sheds light on deep neural
  networks.
\newblock \emph{arXiv preprint arXiv:1901.06523}, 2019.

\bibitem[Yu et~al.(2022)Yu, Lu, Meng, and Karniadakis]{yu2022gradient}
Jeremy Yu, Lu~Lu, Xuhui Meng, and George~Em Karniadakis.
\newblock Gradient-enhanced physics-informed neural networks for forward and
  inverse pde problems.
\newblock \emph{Computer Methods in Applied Mechanics and Engineering},
  393:\penalty0 114823, 2022.

\bibitem[Zabusky and Kruskal(1965)]{zabusky1965interaction}
Norman~J Zabusky and Martin~D Kruskal.
\newblock Interaction of" solitons" in a collisionless plasma and the
  recurrence of initial states.
\newblock \emph{Physical review letters}, 15\penalty0 (6):\penalty0 240, 1965.

\bibitem[Zhao et~al.(2023)Zhao, Ding, and Prakash]{zhao2023pinnsformer}
Zhiyuan Zhao, Xueying Ding, and B~Aditya Prakash.
\newblock {PINN}s{F}ormer: A transformer-based framework for physics-informed
  neural networks.
\newblock \emph{arXiv preprint arXiv:2307.11833}, 2023.

\bibitem[Zhongkai et~al.(2024)Zhongkai, Yao, Su, Su, Wang, Lu, Xia, Zhang, Liu,
  Lu, et~al.]{zhongkai2024pinnacle}
Hao Zhongkai, Jiachen Yao, Chang Su, Hang Su, Ziao Wang, Fanzhi Lu, Zeyu Xia,
  Yichi Zhang, Songming Liu, Lu~Lu, et~al.
\newblock {PINN}acle: A comprehensive benchmark of physics-informed neural
  networks for solving pdes.
\newblock \emph{Advances in Neural Information Processing Systems},
  37:\penalty0 76721--76774, 2024.

\end{thebibliography}

\clearpage
\appendix

\renewcommand{\thesubsection}{\Alph{section}.\arabic{subsection}}
\setcounter{secnumdepth}{2}

\section*{{\LARGE Supplementary Material}}

\section{Exponential Time Differencing}
\label{sec:ap_ETD}

Differential equations typically require numerical methods to obtain approximate solutions, as analytical solutions are seldom available. Among the numerical schemes employed for time-dependent differential equations, Runge–Kutta methods are commonly used due to their conceptual simplicity, ease of implementation, and favorable balance between computational cost and accuracy. Despite these advantages, such methods can perform poorly when applied to stiff systems, where stability constraints may force the time step $\Delta t$ to become prohibitively small, substantially increasing the computational effort required to achieve accurate solutions.

\subsection{Stiffness}

For evolutionary ODE systems in which different components evolve on widely varying timescales, numerical stability constraints are typically dictated by the fastest dynamics \cite{stoer1980introduction}. Hence, the time step required to ensure stability is determined by the faster components, leading to an inefficient use of computational resources when integrating the slower modes. A canonical illustration is provided by a 2D linear system governed by a diagonal vector field with widely separated eigenvalues,
\begin{equation}\label{ODEdiag}
\frac{d}{dt}
\begin{pmatrix}
    u_1 \\
    u_2
\end{pmatrix}
=
\begin{bmatrix}
    -\lambda_{\min} & 0 \\
    0 & -\lambda_{\max}
\end{bmatrix}
\begin{pmatrix}
    u_1 \\
    u_2
\end{pmatrix},
\end{equation}
where$\lambda_{\min}$ and $\lambda_{\max}$ satisfy $0 < \lambda_{\min} \ll \lambda_{\max}$. In this setting, explicit Runge-Kutta schemes are subject to stability restrictions of the form
\begin{equation}
    \Delta t \leq \frac{C}{\|\lambda_{\max}\|}
\end{equation}
for some value of constant $C$, even if the evolution of the slow component $x_1$ alone would permit substantially larger time steps.

An analogous phenomenon arises in higher-dimensional ODE systems and in numerical schemes for PDEs, which can be reduced to a large system of ODEs through spatial discretization by the method of lines or via spectral representation of the solution. For instance, in pseudospectral discretizations of the heat equation (eq.~\eqref{eq:heat1d}, projection in a Fourier basis yields evolution equations of the form
\begin{equation}
\frac{d \hat{u}_n}{d t} = -n^2 \kappa \hat{u}_n\,.
\end{equation}
As the number of spectral modes increases, the highest-frequency components evolve on progressively faster timescales than the low-frequency modes, resulting in increasingly stiff systems and imposing severe stability constraints on explicit time-integration methods.

\subsection{Exponential Time Differencing}\label{sub:etd}

Exponential time differencing methods are designed to address stiffness arising from linear operators by computing the respective part of the propagator in the exact form, usually by evaluating a matrix exponential \cite{cox2002exponential, hochbruck2010exponential}.  Consider a system of ODEs written in the form of eq.\eqref{eq:discretized_PDE}. ETD methods exploit the fact that the linear part $\mathbf{L}$ can be integrated analytically.

Multiplying Eq.~\eqref{eq:discretized_PDE} by the exponential $e^{-\bm{L} t}$ and performing integration along a small interval $\Delta t$ we obtain the following differencing formula
\begin{equation}\label{Edifferencing}
    \hat{\mathbf{u}}(t_{n+1}) \!=\! e^{\Delta t \mathbf{L}} \hat{\mathbf{u}}(t_n) \!+ \!\!\int^{\Delta t}_0 \!\!\!\!e^{\mathbf{L}(\Delta t - s)}\mathbf{N}(\hat{\mathbf{u}}(t_n+s))ds\,,
\end{equation}
where $\Delta t = t_{n+1} - t_n$.
Different approximations of the integral term lead to a family of ETD schemes with different local truncation errors. The simplest approximation assumes that the nonlinear term remains constant over the time step, resulting in the first-order ETD method (ETD-Euler, or ETD-RK1 \cite{cox2002exponential})
\begin{equation}\label{ETDE}
    \hat{\mathbf{u}}_{n+1} = e^{\Delta t\mathbf{L}}\,\hat{\mathbf{u}}_n + \Delta t\,\,\varphi_1(\Delta t\mathbf{L} ) \mathbf{N}(\hat{\mathbf{u}}_n)\,,
\end{equation}
where we define
\begin{align}
    \hat{\mathbf{u}}_n =& \hat{\mathbf{u}}(t_n)\,, \\
    \varphi_1(\mathbf{A})=&\mathbf{A}^{-1}(e^\mathbf{A}-\mathbf{I})\,.
\end{align}
$\varphi_1$ is the first member of a family of $\varphi$-functions that arise in ETD formulations \citep{hochbruck2010exponential}, and which can be described by the following recurrence relation:
\begin{align}
    \varphi_0(\mathbf{A}) =& e^\mathbf{A} \\
    \label{eq:varphi-recurrence} \varphi_{k+1} (\mathbf{A}) =& \mathbf{A}^{-1} \left( \varphi_k(\mathbf{A}) - \frac{1}{k!}\mathbf{I}\right)
\end{align}

Higher order methods can be obtained by introducing midpoints and designing more accurate approximations of the integral in Eq.~\eqref{Edifferencing} via a weighed average of slopes similarly to traditional Runge-Kutta methods. A widely-used fourth-order scheme ETD-RK4 \cite{cox2002exponential} can be defined as
\begin{align}
    & \textbf{a}_n = e^{\frac{\Delta t}{2}\mathbf{L}}\,\hat{\mathbf{u}}_n + \frac{\Delta t}{2} \varphi_1\left(\frac{\Delta t}{2}\mathbf{L}\right)\mathbf{N}(\hat{\mathbf{u}}_n)\,, \\
    & \textbf{b}_n = e^{\frac{\Delta t}{2}\mathbf{L}}\,\hat{\mathbf{u}}_n + \frac{\Delta t}{2} \varphi_1\left(\frac{\Delta t}{2}\mathbf{L}\right)\mathbf{N}(\textbf{a}_n)\,, \\
    & \textbf{c}_n = e^{\frac{\Delta t}{2}\mathbf{L}}\,\textbf{a}_n + \frac{\Delta t}{2} \varphi_1\left(\frac{\Delta t}{2}\mathbf{L}\right)(2\mathbf{N}(\textbf{b}_n) - \mathbf{N}(\hat{\mathbf{u}}_n))\,,\\
    &\!\! \mathbf{\Phi}(\hat{\mathbf{u}}_n) \!=\!    \textbf{B}_1 \mathbf{N}(\hat{\mathbf{u}}_n)
    \!+\! \textbf{B}_2 \mathbf{N}(\mathbf{a}_n)
  \!+\! \textbf{B}_3 \mathbf{N}(\mathbf{b}_n)
    \!+\! \textbf{B}_4 \mathbf{N}(\mathbf{c}_n)\,, \\
    & \hat{\mathbf{u}}_{n+1}
= e^{\Delta t \mathbf{L}} \hat{\mathbf{u}}_n
  + \Delta t \,\mathbf{\Phi}(\hat{\mathbf{u}}_n)\,, 
  \label{eq:ETD-step}
\end{align}
where coefficient matrices $\textbf{B}_j$ are defined in terms of the high-order $\varphi$-functions as
\begin{align}
    & \textbf{B}_1 = \varphi_1(\Delta t\mathbf{L}) - 3 \varphi_2(\Delta t\mathbf{L}) + 4 \varphi_3(\Delta t\mathbf{L})\,, \\
    & \textbf{B}_2 = \textbf{B}_3 = 2 \big(\varphi_2(\Delta t\mathbf{L})-2\varphi_3(\Delta t\mathbf{L})\big)\,, \\
    & \textbf{B}_4 = -\varphi_2(\Delta t\mathbf{L}) + 4 \varphi_3(\Delta t\mathbf{L})\,.
\end{align}
By treating the stiff linear dynamics exactly while retaining explicit evaluations of the nonlinear terms, ETD methods significantly relax stability constraints on the time step. This property makes them particularly well-suited for ODE systems obtained by spectral discretizations of partial differential equations, where stiffness often originates in the linear operator.

\subsection{Details on the stiffness of the ODEs emerging from PDE discretization}
\label{sec:ap_stiff}

In this subsection we discuss the stiffness of ODEs emerging from the spectral representation of a PDE solution. For simplicity we assume that spatial coordinate $x$ is 1-dimensional.

If the operator $\mathcal{L}$ in Eq.~\eqref{eq:pde_split} involves spatial derivatives up to order $m$ with respect to the $x$ then its Fourier-domain counterpart is a diagonal matrix that has the entries of the form of a constant-coefficient polynomial $P_m(k)$, where $k$ is the variable indexing the spectral basis functions, e.g., $\sin(\pm 2\pi  k x)$ or $\cos(\pm 2\pi  k x)$ (this is the basis used in our experiments). Assuming that the total number of basis elements is $K$, we can show that largest element of the matrix $\bm{L}$ is of the order $\lambda_{\max}= \mathcal{O}(K^p)$, and the smallest one is $\lambda_{\min}= \mathcal{O}(1)$, where $p=m$ for parabolic equations and $p=m/2$ for second-order hyperbolic equations.
Since $\lambda_{\max}/\lambda_{\min} = \mathcal{O}(K^p) \gg 1$, the semi-discrete system \eqref{eq:discretized_PDE} obtained by spectral representation of PDE \eqref{eq:pde_split} exhibits widely separated time scales. 

For parabolic equations, large $k$ correspond to rapidly decaying modes, while for hyperbolic equations they reflect fast oscillatory dynamics. In both cases, explicit time-integration schemes must adopt time steps constrained by the largest $k$, leading to prohibitively small time steps and stiff or stiff-like behavior in the induced ODE system \cite{canuto2007spectral, leveque2007finite, ortega1990_numerical_analysis_second_course, burden2016_numerical_analysis}.

\section{Experiment Details}
\label{sec:ap_exp_setup}

For all experiments, the architecture hyperparameters in the baseline models were set according to \citet{Bizzi2025NeuroSpectral}, and the corresponding configurations for \edpmethod{} were chosen so as to keep a similar parameter count.

Other hyperparameters were set after preliminary tests in order to achieve the best performance for each model under the same order of magnitude for training time. The hyperparameter values are reported in Table~\ref{tab:forward-hyperparam}.

Since the Neural ODE relies on numerical integration, machine precision is important, therefore for all experiments involving \edpmethod{} we use a 64-bit representation for floating point numbers. For the MLP baselines, however, it was verified experimentally that the added precision of \texttt{float64} does not improve results, but rather only makes training and inference slower, hence a 32-bit representation was used for them.

For all experiments, an exponential learning rate scheduler was used, with decay factor $\gamma=0.995$ for \pdemethod and $\gamma=0.999$ for the baselines. These values were chosen after thorough testing to achieve the best performance for each method. 

\begin{table*}[htb]
\caption{Learning rate, number of training steps and architecture shape used in each of the forward experiments.}
\label{tab:forward-hyperparam}
\centering
\begin{tabular}{@{}lcccccc@{}}
\toprule
\multirow{2}{*}{} & \multicolumn{4}{c}{\textbf{\pdemethod{}}}      & \multicolumn{2}{c}{Baselines} \\ \cmidrule(l){2-5} \cmidrule(l){6-7} 
                     & LR   & N. Steps & Width & Depth & LR            & N. Steps        \\ \midrule
\multicolumn{1}{l|}{Heat 1D}           & 0.02 & 500    & 796   & \multicolumn{1}{c|}{2} & 0.001         & 5000          \\
\multicolumn{1}{l|}{Burgers}           & 0.02 & 500    & 800   & \multicolumn{1}{c|}{2} & 0.001         & 5000          \\
\multicolumn{1}{l|}{Wave}              & 0.02 & 1000   & 300   & \multicolumn{1}{c|}{2} & 0.001         & 5000          \\
\multicolumn{1}{l|}{Korteweg--De Vries} & 0.01 & 1000   & 400   & \multicolumn{1}{c|}{2} & 0.001         & 10000         \\
\multicolumn{1}{l|}{Heat 2D}           & 0.01 & 500    & 100   & \multicolumn{1}{c|}{2} & 0.001         & 20000         \\ \bottomrule
\end{tabular}
\end{table*}

\subsection{Evaluation metric}

All experiments were evaluated on the \textbf{relative root mean squared error}, as in \citet{wu2024ropinn}:
\begin{equation}
    \text{rRMSE} = \sqrt{ \frac{ \sum_{i=1}^N \left( {u_{\mathrm{pred}}}(t_i,\mathbf{x}_i) - {u_{\mathrm{ref}}}(t_i,\mathbf{x}_i) \right)^2} {\sum_{i=1}^N \left( {u_{\mathrm{ref}}}(t_i,\mathbf{x}_i)\right)^2}}\;.
\end{equation}

\subsection{Additional results}

In addition to the mean value for training time and rRMSE presented in Table~\ref{table:results}, we also report the standard deviation of these quantities in Table~\ref{table:stds}.

\begin{table*}[h!]
\centering
\caption{\textbf{Quantitative evaluation on PDE benchmarks}. Below we list the standard deviation for the relative root mean square error (rRMSE) and training time (TT, in seconds) for all forward problems, across 10 different runs with different random seeds.}
\setlength{\tabcolsep}{6pt}
\begin{tabular}{@{}lcccccccccc@{}}
\toprule
\multirow{2}{*}{\textbf{Model}} & \multicolumn{2}{c}{\textbf{Heat 1d}}                 & \multicolumn{2}{c}{\textbf{Burgers 1d}}              & \multicolumn{2}{c}{\textbf{Wave 1d}}                 & \multicolumn{2}{c}{\textbf{KdV 1d}}                  & \multicolumn{2}{c}{\textbf{Heat 2d}} \\ \cmidrule(l){2-11} 
                                & \textbf{rRMSE}  & \multicolumn{1}{c|}{\textbf{TT}}   & \textbf{rRMSE}  & \multicolumn{1}{c|}{\textbf{TT}}   & \textbf{rRMSE}  & \multicolumn{1}{c|}{\textbf{TT}}   & \textbf{rRMSE}  & \multicolumn{1}{c|}{\textbf{TT}}   & \textbf{rRMSE}     & \textbf{TT}     \\ \midrule
PINN                            & 0.0049          & \multicolumn{1}{c|}{\textbf{0.02}} & 0.0187          & \multicolumn{1}{c|}{\textbf{0.55}} & 0.0682          & \multicolumn{1}{c|}{3.48}          & 0.0552          & \multicolumn{1}{c|}{\textbf{0.32}} & 0.0125             & 0.24            \\
QRes                            & 0.0050          & \multicolumn{1}{c|}{0.04}          & 0.0098          & \multicolumn{1}{c|}{1.22}          & 0.0137          & \multicolumn{1}{c|}{6.67}          & 0.0210          & \multicolumn{1}{c|}{0.40}          & 0.0074             & 0.61            \\
FLS                             & 0.0457          & \multicolumn{1}{c|}{0.04}          & 0.0249          & \multicolumn{1}{c|}{0.64}          & 0.0463          & \multicolumn{1}{c|}{1.57}          & 0.0572          & \multicolumn{1}{c|}{0.33}          & 0.0139             & 0.59            \\
\textbf{\pdemethod{}}           & \textbf{0.0008} & \multicolumn{1}{c|}{0.23}          & \textbf{0.0013} & \multicolumn{1}{c|}{0.97}          & \textbf{0.0004} & \multicolumn{1}{c|}{\textbf{1.09}} & \textbf{0.0004} & \multicolumn{1}{c|}{2.24}          & \textbf{0.0009}    & \textbf{0.07}   \\ \bottomrule
\end{tabular}
\label{table:stds}
\end{table*}

\subsection{1D Heat equation}

In this experiment, the nonlinear vector field was represented by a multilayer perceptron having two hidden layers with standard LeakyReLU activation. The number of neurons in each hidden layer is four times the size of the input (which corresponds to the adopted number of spatial frequencies). 

For \eqref{eq:heat1d} we used Gaussian initial condition of the form
\begin{align}
    u(0,x) = \frac{1}{\sqrt{2\pi}}\exp\left(\frac{-(x-0.4)^2}{2\sigma^2}\right)\,, \mbox{ where } \sigma = 0.5\,. 
\end{align}

We define the linear field $\mathcal{L}$ to be the homogeneous version of eq.~\eqref{eq:heat1d}, with $\kappa(x)=\kappa_0=0.1$, i.e. 
\begin{equation}
    \mathcal{L} u = \kappa_0 \frac{\partial^2 u}{\partial x^2}
\end{equation}

In the sine basis, which we adopt because it automatically enforces the homogeneous Dirichlet boundary conditions, this field is a diagonal linear map, given by
\begin{equation}
    (\mathbf{L} \hat{\mathbf{u}})_i = -\kappa_0 \omega_i^2\hat{u}_i\,.
\end{equation}

\begin{figure}
    \centering
    \includegraphics[scale=0.6]{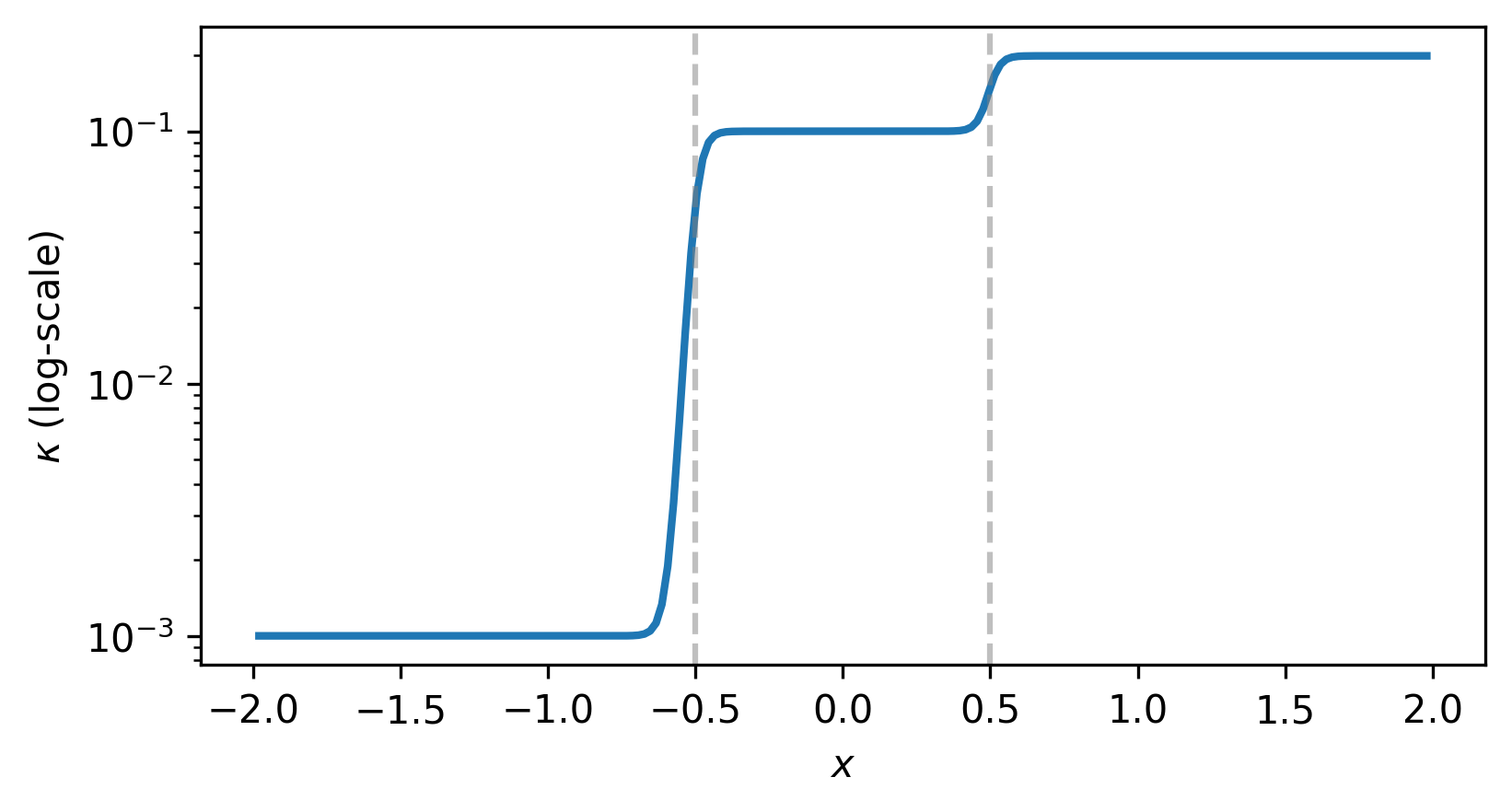}
    \caption{The conductivity profile used in the experiment with one-dimensional heat equation \eqref{eq:heat1d}.}
    \label{fig:heat1d_conductivity}
\end{figure}

The reference solution for the one-dimensional heat equation was obtained by a pseudo-spectral method using a standard fourth-order Runge-Kutta method, with $\Delta x = 2\times10^{-2}$ and $\Delta t = 2\times10^{-4}$.

\edpmethod{} was trained with 199 frequencies, which corresponds to 201 collocation points in space including the boundaries, where the solution is identically zero, and 201 time steps (leading to a $0.01$ time step).
The baseline models were trained on a $201\times201$ grid. 

\subsection{Viscous Burgers' equation}

The architecture of the multilayer perceptron used here to represent the nonlinear term is precisely the same as in the case of the one-dimensional heat equation.

Because of the periodic boundary conditions and the presence of odd-order derivatives, we used a Fourier basis for the pseudo-spectral decomposition. The natural candidate for the linear field $\mathcal{L}$ is 
\begin{equation}
    \mathcal{L} u = \nu \frac{\partial^2u}{\partial x^2}\,,
\end{equation}
which is diagonal in the Fourier basis, similar to the heat equation:
\begin{equation}
    (\mathbf{L} \hat{\mathbf{u}})_i = -\nu \omega_i^2 \hat{u}_i\,.
\end{equation}

The reference solution for the Burgers equation was generated via a pseudo-spectral method, advanced in time using a fourth-order Runge--Kutta integrator, with spatial and temporal step sizes $\Delta x = 10^{-2}$ and $\Delta t = 5\times10^{-5}$, respectively.

The \edpmethod{} model was trained with $200$ frequencies and $401$ time steps (leading to a $2.5\times 10^{-3}$ time step), while the baseline models were trained in a $401\times201$ grid.

\subsection{1D Wave equation}

\begin{figure}[hb]
    \centering
    \includegraphics[width=0.8\linewidth]{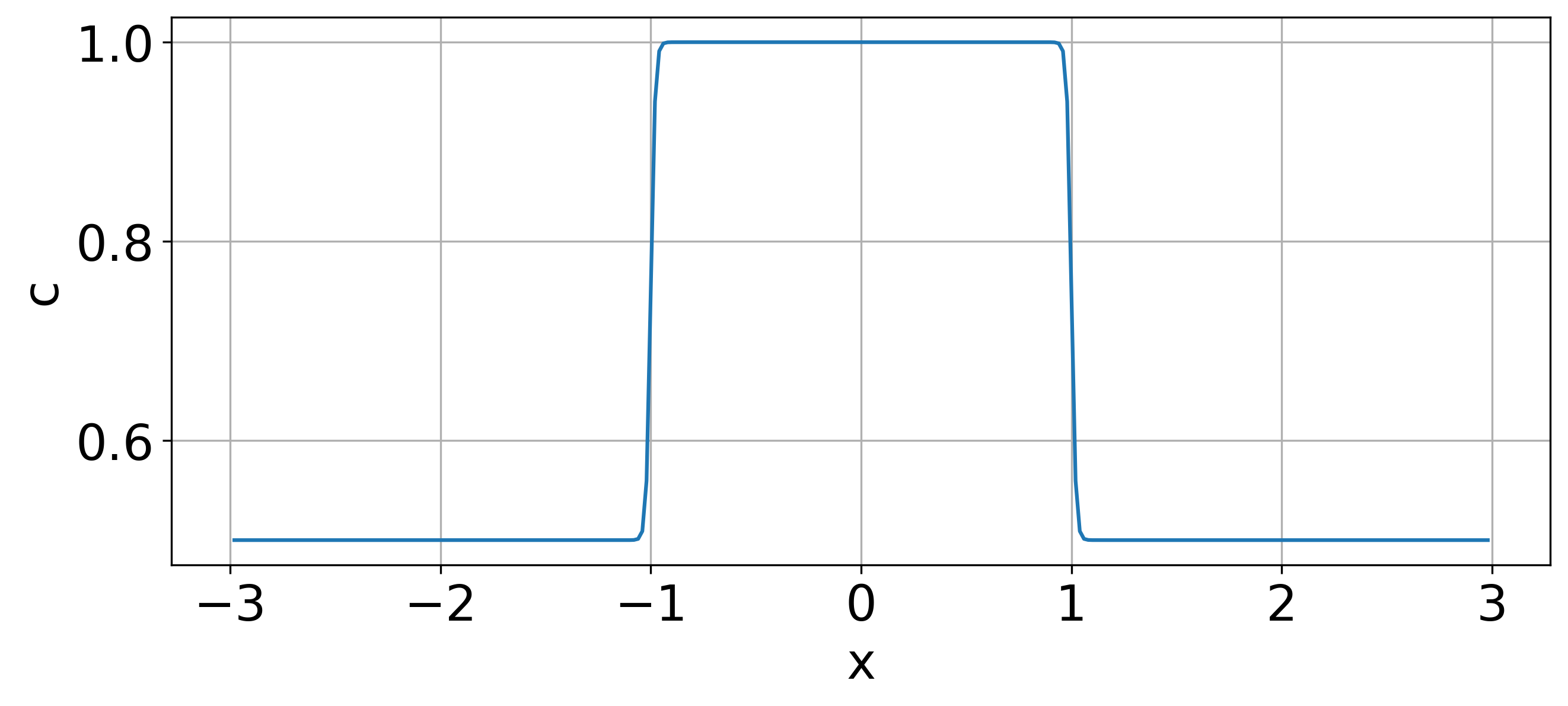}
    \caption{Wave propagation velocity $c(x)$ for the wave equation's simulation domain.}
    \label{fig:wave_config}
\end{figure}

The wave propagation domain in our experiment is a three-layered medium with a higher-velocity "bump" in the middle, and lower velocity near the edges, with the interfaces smoothed by sigmoid functions. The profile of the velocity $c(x)$ is presented in Fig.~\ref{fig:wave_config}.

The initial condition in this experiment is
\begin{align}
    u(0,x) =&
    \frac{1}
         {\sqrt{2\pi}\sigma}\exp\left(\frac{-x^2}{2\sigma^2}\right)\,, \\
         u_t (0, x) =& 0\,,
\end{align}
for $\sigma=0.2$.

Since since this problem has a second order temporal derivative, we have to turn it into a first-order system by writing the state function as $[u, v]$, where $v$ stands for the first derivative of $u$, and so the true vector field is written as
\begin{equation}
    \mathcal{F}\begin{pmatrix}
        u \\ v
    \end{pmatrix} = \begin{pmatrix}
        v \\ c^2 \frac{\partial^2 u}{\partial x^2}
    \end{pmatrix} \,.
\end{equation}

The linear field $\mathcal{L}$ is defined as the vector field for a homogeneous wave problem with $c_0=1$:

\begin{equation}
    \mathcal{L} \begin{pmatrix}
        u \\ v
    \end{pmatrix}
    =  \begin{pmatrix}
    v \\
    c_0^2\frac{\partial^2 u}{\partial x^2}
    \end{pmatrix} \,,
\end{equation}

which in the sine basis becomes
\begin{equation} \label{eq:wave_L}
    \left[\mathbf{L} \begin{pmatrix}    
    \hat{\mathbf{u}} \\ \hat{{\mathbf{v}}}  \end{pmatrix}
    \right]_i     =  \begin{pmatrix}
    -\hat{v}_i \\ c_0^2 \omega_i^2 \hat{u}_i
    \end{pmatrix} \,.
\end{equation}
This linear operator is a diagonal linear map $\mathbf{L}: (\mathbb{R}^2)^N \mapsto(\mathbb{R}^2)^N$, and thus its exponential involves the exponentiation of $2\times 2$ matrices. More details on this calculation can be found in Appendix \ref{sec:ap_implementation_details}.

For \edpmethod{}, the nonlinear part of the vector field is represented by an MLP with 2 hidden layers of 300 neurons each, with a LeakyReLU activation function. We use the sine basis with 299 frequencies, which automatically enforces Dirichlet boundary conditions, and integrate for 101 timesteps of $\Delta t=2\times 10^{-2}$.

In the MLP baseline solutions we use the same collocation points in space, but 201 points in time, with relative weights of $10^3$ and $10^2$ for the initial and boundary conditions, as MLPs tend to struggle with learning them.
The reference solution for this problem was obtained by a pseudospectral method with $\Delta x=2 \times 10^{-2}$, using a 4th-order Runge-Kutta solver with $\Delta t=10^{-3}$.

\subsection{Korteweg--De Vries equation}

For this equation, the nonlinear vector field is modeled by a multilayer perceptron with the same architecture used for the Burgers and one-dimensional heat equations.

Periodic boundary conditions are imposed as hard constraints by adopting a Fourier basis, as in Burgers' equation. 
In this basis, the linear field for KdV is written as
\begin{align}
    \mathcal{L} u =& \delta \frac{\partial^3}{\partial x^3} u \,, \\
    \label{eq:kdv-L} \mathbf{L} \begin{pmatrix}    
    \hat{u}^{(c)}_i \\ \hat{{u}}^{(s)}_i  \end{pmatrix} =& \delta \omega_i^3 \begin{pmatrix}
        -\hat{{u}}^{(s)}_i \\ \hat{{u}}^{(c)}_i
    \end{pmatrix} \,.
\end{align}
As in the wave equation, the exponential of this operator becomes the exponential of a $2\times 2$ matrix, which is detailed in Appendix~\ref{sec:ap_implementation_details}.

The reference solution for the Korteweg--De Vries equation was computed using a pseudo-spectral approach combined with a classical fourth-order Runge--Kutta time integrator, with spatial and temporal step sizes $\Delta x = 2\times 10^{-2}$ and $\Delta t = 10^{-5}$, respectively.

The \edpmethod{} model was trained with $100$ spatial frequencies and $501$ time steps, and the baseline models were trained on a $501\times101$ grid.

 \subsection{2D Heat equation}

In the two-dimensional heat equation, the nonlinear vector field for \pdemethod{} is represented by a multilayer perceptron with transpose layers in between standard hidden layers \cite{Bizzi2025NeuroSpectral}, since the input is not a vector, but a matrix of spectral coefficients. We train the model with $2$ such hidden layers of width $100$ each.

The initial condition considered is
\begin{align}
    u(0,x) = \left[1-\frac{1}{\psi(x)}\right]\left[1-\frac{1}{\psi(y)}\right],
\end{align}
where
\begin{align}
\psi(z) = 1+\exp(-50(|z|-0.5)).
\end{align}

As in the one-dimensional case of the heat equation, homogeneous Dirichlet boundary conditions suggest the adoption of a sine basis for the spectral decomposition. Again, the linear operator of the second spatial derivative is diagonal and invertible, thus we may implement our ETD integrator via Hadamard (elementwise) products. 

\begin{figure}
\centering
\includegraphics[scale=0.5]{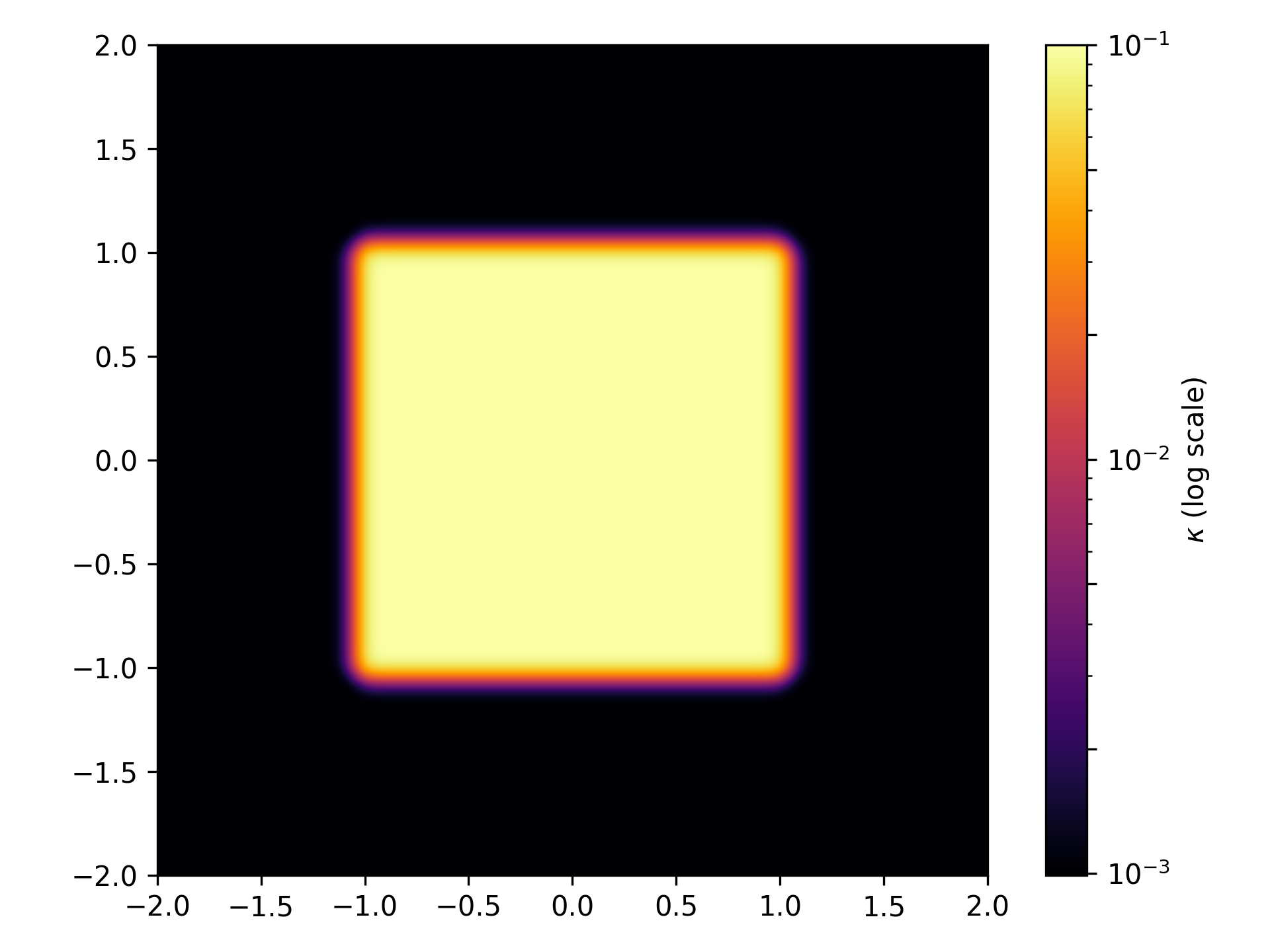}
\caption{The conductivity profile considered in the 2d heat equation.}
\label{fig:conductivity}
\end{figure}

\begin{figure*}
\centering
\includegraphics[width=0.75\linewidth]{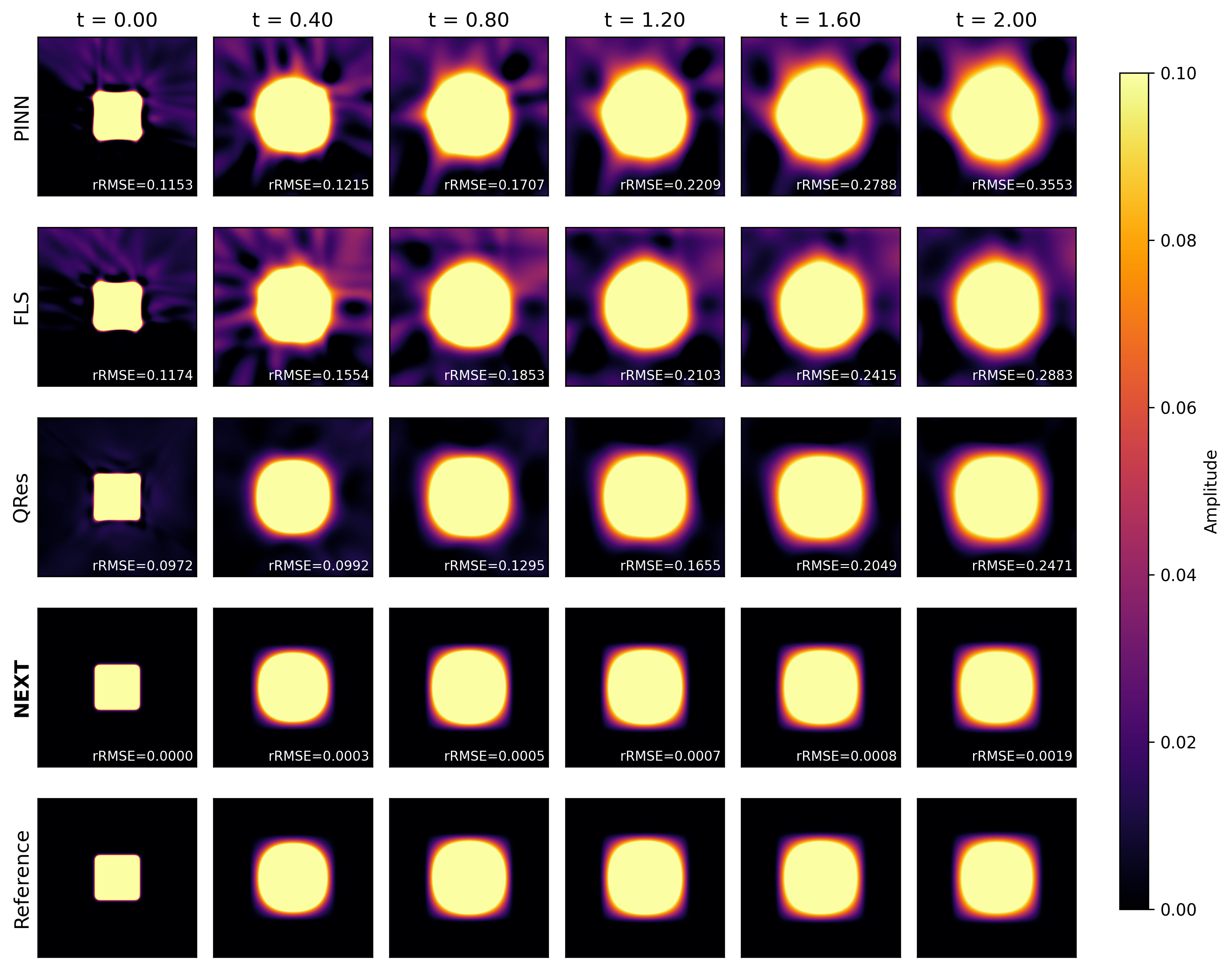}
\caption{Predicted solutions for the 2d heat equation. The rRMSE is computed for each time snapshot. \pdemethod{} correctly predicts the solution, while the baseline models have difficulty dealing with the sharp interfaces between materials.}
\label{fig:heat2d}
\end{figure*}

\begin{figure*}
\centering
\includegraphics[width=0.75\linewidth]{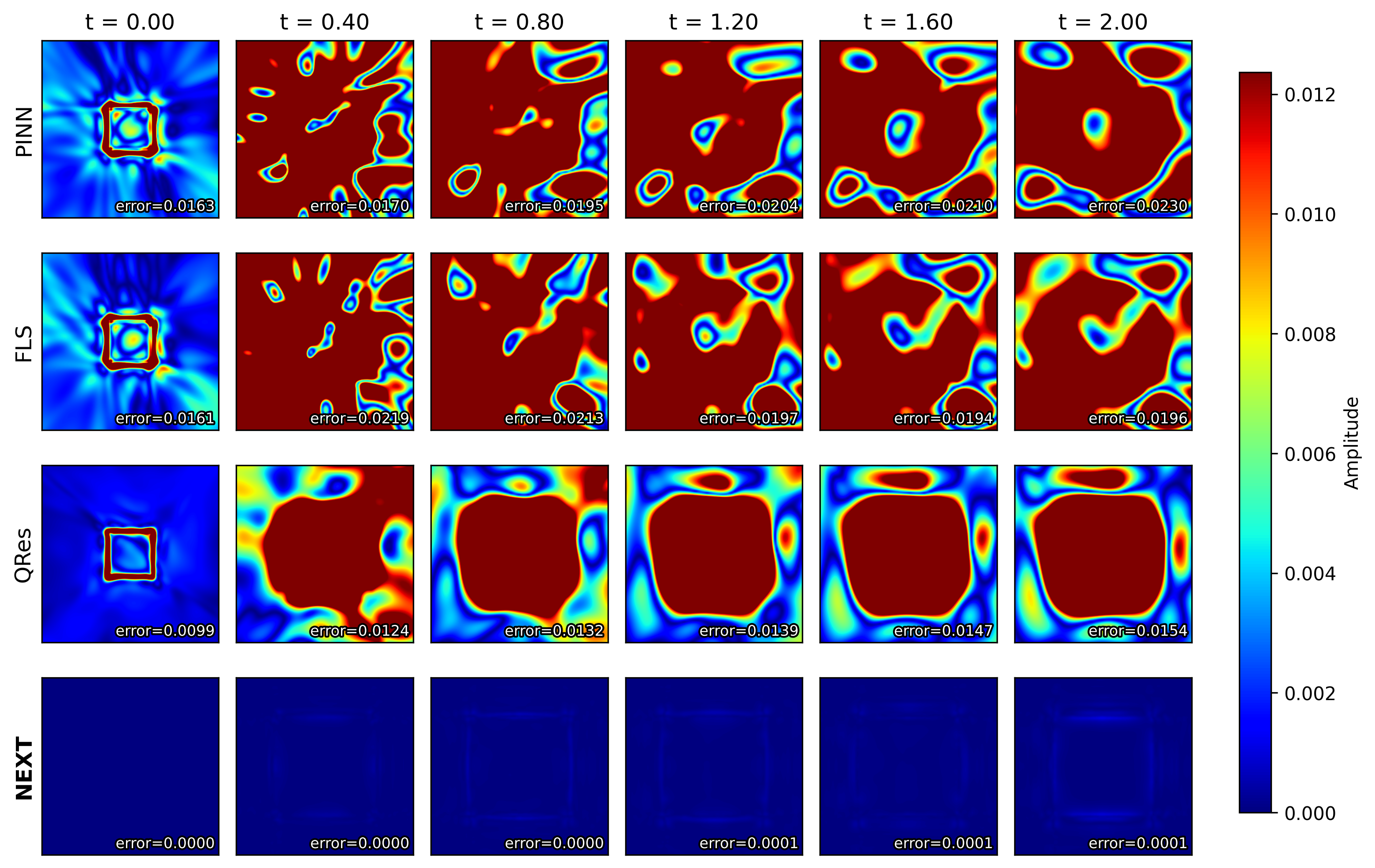}
\caption{Absolute error of each model prediction for the two-dimensional heat equation. The displayed error is the mean absolute error for each time snapshot. The baseline models fail to learn the initial condition, and this error propagates in time, leading to a very inaccurate solution at later times. \pdemethod{}, on the other hand, has a more stable error profile, due to its causal nature and to its capacity to automatic enforcement of initial conditions.}
\label{fig:heat2d_error}
\end{figure*}

Fig.~\ref{fig:conductivity} displays the conductivity profile, and in Fig.~\ref{fig:heat2d} we plot the comparison between \pdemethod{} and all the baseline models. The absolute error of each model is shown in Fig.~\ref{fig:heat2d_error}. 

The reference solution for the two-dimensional heat equation was obtained via a pseudo-spectral method combined with a fourth-order Runge--Kutta time integrator, using $\Delta x = \Delta y = 4\times10^{-2}$ and $\Delta t = 4\times10^{-4}$.

The \edpmethod{} model was trained with $99$ frequencies in each spatial dimension, and $201$ time steps. The adoption of a sine basis imposes the homogeneous Dirichlet condition as a hard constraint. Because it is unfeasible, in terms of memory, to train the baseline models in the complete grid, in each training step we sampled $10000$ points to enforce the equation residue, $1000$ points for the initial condition and $500$ points for the boundary condition. 

\subsection{Burgers parameter identification}
 
For this experiment, \edpmethod{} was trained for 500 Adam steps with learning rate 0.01, while the MLP baselines were trained for 5000 steps with learning rate 0.001. In all cases, the data-driven loss had weight 10 and the physics-informed loss had weight 1, which was observed to improve results in preliminary experiments.

\subsection{Boundary heat flux identification}

Due to the different scales of the problem parameters, we normalized them before training. The \edpmethod{} was trained for 5000 steps with learning rate of 0.02 with the optimizers Adam, on a 101×51 grid, and we adopt the Fourier basis because the boundary constraint should be less strict. The data-driven loss and the physics-informed loss had the same weight. To test the method's robustness, we ran 10 different seeds.  To accommodate a transient boundary condition, we extended the spatial domain by doubling the length $L_x$, while evaluating the physics loss only within the original domain. The physical parameters used are shown in Table~\ref{table:Parameters}.

\begin{table}[]
\centering
\begin{tabular}{@{}lll@{}}
\toprule
\textbf{Parameters} & \textbf{Unit} & \textbf{Value} \\ \midrule
$\rho$        & kg/m$^3$              & 3120    \\
$c_p$         & J/(kg·K)           & 850     \\
$k_0$         & W/(m·K)            & 4.86    \\
$k_1$         & W/(m·K$^2$)        & 0.0016  \\
$h$           & W/(m$^2$·K)           & 7       \\
$\sigma$      & W/(m$^2$·K$^4$)    & 5.67e-08 \\
$\varepsilon$ & —                  & 1       \\
$T_0$         & K                  & 300     \\
$T_a$         & K                  & 300     \\ \bottomrule
\end{tabular}
\caption{Thermophysical parameters for eq.~\eqref{eq:governing}}
\label{table:Parameters}
\end{table}

\section{Ablation Experiments}
\label{sec:ap_experiments}

\subsection{Order of the ETD solver}

We perform an ablation experiment on the effect of using a high-order vs low-order exponential solver with \pdemethod{}, and show the results for the 1D wave and 2D heat equations in Table~\ref{tab:solver-ablation}. 

First-order exponential integration methods have been shown to obtain good results on Neural ODEs for stiff problems \citep{fronk2025training}.
The fourth-order integrator used in \pdemethod{}, however, is significantly more accurate than the first-order method, albeit at a larger computational cost. We also trained the model with fourth-order Runge-Kutta (equivalent to NeuSA), but it diverged numerically in all experiments due to stiffness.

\begin{table}[htb]
\centering
\caption{Relative root mean square error and training time (in seconds) obtained by \pdemethod{} using different orders of exponential integrator.}
\label{tab:solver-ablation}
\begin{tabular}{@{}lcccc@{}}
\toprule
{\multirow{2}{*}{\textbf{Solver}}} & \multicolumn{2}{c}{\textbf{Wave 1D}} & \multicolumn{2}{c}{\textbf{Heat 2D}} \\
\multicolumn{1}{c}{}                        & \textbf{rRMSE}         & \textbf{TT}          & \textbf{rRMSE}         & \textbf{TT}          \\ \midrule
\textbf{ETD1}                                        & 0.0629        & 38.6        & 0.0031        & 66.2        \\
\textbf{ETD4}                                        & 0.0030        & 130.0       & 0.0007        & 215.7       \\ \bottomrule
\end{tabular}
\end{table}

\subsection{Architecture ablation}

We train \pdemethod{} on the 2D heat equation with different width and depth configurations under the same random seed, and show the results in Table~\ref{tab:architecture-ablation}. Accuracy is shown to be relatively robust to changes in architecture, although the training time increases with the number of parameters. 

{\setlength{\tabcolsep}{6pt}
\begin{table}[htbp]
\centering
\caption{\pdemethod{}'s performance in the 2D Heat equation by Width (W) and Depth (D) of the neural vector field. A dash (-) indicates training did not converge. The architecture chosen for the experiment reported in the paper (underlined below) was the one that achieved the best trade-off between accuracy and training time.}
\label{tab:architecture-ablation}
\begin{tabular}{@{}l ccc ccc@{}}
\toprule
                              & \multicolumn{3}{c}{\textbf{rRMSE}}   & \multicolumn{3}{c}{\textbf{TT}}      \\ \cmidrule(lr){2-4} \cmidrule(lr){5-7} 
\textbf{W \textbackslash \ D} & \textbf{1} & \textbf{2} & \textbf{3} & \textbf{1} & \textbf{2} & \textbf{3} \\ \midrule
\textbf{100}                           & 0.0015     & \underline{0.0008}     & 0.0019     & 161        & \underline{236}        & 313        \\
\textbf{200}                           & 0.0015     & 0.0006     & 0.0013     & 214        & 344        & 473        \\
\textbf{300}                           & 0.0031     & 0.0011     & -          & 264        & 449        & -          \\
\textbf{400}                           & 0.0037     & 0.0025     & 0.0029     & 333        & 804        & 1274       \\ \bottomrule
\end{tabular}
\end{table}}

\section{Implementation Details}
\label{sec:ap_implementation_details}

\subsection{$\mathbf{L}$ and $\varphi$ in PDEs}
\label{subsec:linsep}

We can represent an approximation of the solution to a 1D time-evolving PDE in a spectral basis using a vector ${\hat{\mathbf{u}}} \in \mathbb{R}^N$, in the form
\begin{equation}
    u(t, x) = \sum^{N-1}_{n=0} \hat{{u}}_n(t)b_n(x), 
\end{equation}
which, if evaluated on a vector of collocation points $\mathbf{x}=(x_1,\dots x_N)^T$, becomes the ODE
\begin{equation}
\label{eq:PDE_hat}
        \frac{d\hat{\mathbf{u}}}{dt} = F(t,\mathbf{x},\hat{\mathbf{u}},\nabla \hat{\mathbf{u}}, \nabla^2\mathbf{\hat{u}},\dots),
\end{equation}
where the space derivatives of $\hat{\mathbf{u}}$ are calculated using the basis $\mathbf{b_1}, \dots, \mathbf{b_N}$. If the chosen basis is a set of eigenfunctions of some of the operators in the PDE (e.g. sine or cosine are eigenfunctions of the Laplacian), then the linear part $\mathbf{L}$ of the field becomes diagonal:
\begin{equation}\label{eq:F_separation}
\begin{aligned}
    F(\hat{\bm{u}}) =  \mathbf{L}\hat{\mathbf{u}}+ N(\hat{\bm{u}}), \\
    \text{ where } (\mathbf{L}\hat{\mathbf{u}})_n =  \lambda_n \hat{u}_n .
\end{aligned}
\end{equation}

This diagonalization turns the exponentiation of $\mathbf{L}$ and its multiplication with $\hat{\mathbf{u}}$ into element-wise operations, which is what makes the Exponential Time Differencing algorithm efficient.

Some consideration must also be made for the $\bm{\varphi}$-functions used in ETD. An important linear operator in PDEs is the Laplacian, which has $\cos(\omega\textbf{x})$ and $\sin(\omega\textbf{x})$ as eigenfunctions, hence their wide use as bases in classical pseudospectral methods and in our decompositions.

In problems with Dirichlet boundary conditions, we use the sine basis, where the state of the system is defined by
\begin{equation}
    u(x) = \sum_{n=1}^{N} \hat{u}_n \sin(\omega_n x),
\end{equation}
and thus
\begin{equation}
    \nabla^2\hat{\mathbf{u}} = \sum_{n=1}^{N} -\omega_n^2\hat{u}_n + \mathbf{N}(\hat{\mathbf{u}}),
\end{equation}
which means $\mathbf{L}$ is given by a diagonal matrix
\begin{equation}
    L = 
    \begin{bmatrix}
        -\omega_1^2 & & 0\\
         & \ddots & \\
         0 & & -\omega_N^2
    \end{bmatrix} \,.
\end{equation}

In problems where a first derivative is needed or periodic boundary conditions are imposed, one must use the Fourier basis:
\begin{equation}
    u(x) = \sum_{i=0}^{N_c} u_i^{(c)}\cos (\omega_i x) + \sum_{i=1}^{N_s} u_i^{(s)}\sin (\omega_i x) \,,
\end{equation}
where we have a generalized coefficient vector as
\begin{equation}
    \hat{\textbf{u}} = 
    \begin{pmatrix}
        \textbf{u}^{(c)} \\
        \textbf{u}^{(s)}
    \end{pmatrix}.
\end{equation}

The Laplacian operator in Fourier basis is also diagonal, as 
\begin{align}
    \nabla^2\cos{(\omega x)} =& -n^2 \cos{(\omega x)} \\
    \nabla^2\sin{(\omega x)} =& -n^2 \sin{(\omega x)},
\end{align}
However, the cosine term for $n=0$ is a constant, and so its Laplacian is zero, meaning that, in order to calculate the $\varphi$ functions for this coordinate, we must take the limit, for instance,
\begin{equation}
    \varphi_{1}(0) = \lim_{z\rightarrow0}\frac{e^z - 1}{z} = 1 \,.
\end{equation}

\subsection{PDEs with second derivatives}

In PDEs such as the wave equation, there is a second time-derivative, as in
\begin{equation}
    \frac{d^2\hat{\mathbf{u}}}{dt^2} = \tilde L \left(\hat{\mathbf{u}}, {\frac{d\hat{\mathbf{u}}}{dt}}\right) + \textbf{N}\left(\hat{\mathbf{u}}, {\frac{d\hat{\mathbf{u}}}{dt}}\right)\,.
\end{equation}
By writing the state vector as $[\hat{\mathbf{u}}, \hat{\mathbf{v}}]$, where $\hat{\mathbf{v}}$ stands for the time-derivative of $\hat{\mathbf{u}}$, the linear operator in eq.~\eqref{eq:wave_L} then acts independently on the $n$-th frequency mode as a $2\times 2$ matrix:
\begin{equation}
    \mathbf{L}_n = \begin{bmatrix}
        0   &   1 \\
        -c_0^2\omega_n^2    &   0
    \end{bmatrix} \,.
\end{equation}

The exponential of $h\mathbf{L}$, which is required for ETD, will be simply the concatenation of the exponential for each $h\mathbf{L}_n$ matrix, which has a closed form. Writing $\rho = c_0\omega_n$, we have
\begin{equation}
    \varphi_0(h\mathbf{L}_n) = e^{h \mathbf{L}_n} = \begin{bmatrix}
        \cos(h \rho)          &         \frac{\sin(h \rho)}{\rho}          \\
        -\rho \sin{(h \rho)}    &      \cos(h \rho)
    \end{bmatrix} \,,
\end{equation}
and the recurrence relation in eq.~\eqref{eq:varphi-recurrence} allows one to obtain the $\varphi$ terms as well:
\begin{align}
    \varphi_1(h\mathbf{L_n}) =& \begin{bmatrix}
        \frac{\sin(h \rho)}{h \rho}     &   \frac{1 - \cos(h \rho)}{h \rho^2} \\
        \frac{\cos(h \rho) - 1}{h}  &   \frac{\sin(h \rho)}{h \rho}
    \end{bmatrix} \,, \\
    \varphi_2(h\mathbf{L_n}) =& \begin{bmatrix}
        \frac{1-\cos(h\rho)}{h^2 \rho^2}     &   \frac{h \rho - \sin(h \rho)}{h^2 \rho^3} \\
        \frac{\sin(h \rho)-h \rho}{h^2 \rho}  &   \frac{1-\cos(h\rho)}{h^2 \rho^2}
    \end{bmatrix} \,, \\
    \varphi_3(h\mathbf{L_n}) =& \begin{bmatrix}
        \frac{h \rho - \sin(h \rho)}{h^3 \rho^3}     &   \frac{\cos(h\rho) - 1 + \frac{1}{2}h^2\rho^2}{h^3 \rho^4} \\
        \frac{\cos(h\rho) - 1 + \frac{1}{2}h^2\rho^2}{h^3 \rho^2}  &   \frac{h \rho - \sin(h \rho)}{h^3 \rho^3}
    \end{bmatrix} \,.
\end{align}

Calculating these matrices requires only the computation of elementwise sines and cosines, which makes it particularly fast.

\subsection{Odd derivatives in the Fourier basis}

In the KdV equation, the linear field consists of the third space derivative of $u$. In the Fourier basis used here, the operator $\mathbf{L}$ won't be diagonal, as the derivatives of the sine modes depend on the cosine modes and vice-versa, as in eq.~\eqref{eq:kdv-L}. However, it still acts independently on each of the 2-vectors given by $[\hat{u}^{(c)}_n, \hat{u}^{(s)}_n]$. In this sense, the linear operator corresponding to a space derivative of order $2p+1$ is caracterized by the $2\times 2$ matrices
\begin{equation}
    \mathbf{L}_n = \omega_n^{2p+1}\begin{bmatrix}
        0 & -1\\
        1 & 0
    \end{bmatrix}
\end{equation}

The $\varphi$ functions for this operator are analogous to those for the wave equation, and can be calculated in an analogous manner:
\begin{align}
    \varphi_0(h\mathbf{L}_n) =& e^{h \mathbf{L}_n} = \begin{bmatrix}
        \cos\gamma          &         \sin\gamma          \\
        -\sin\gamma    &      \cos\gamma
    \end{bmatrix} \,, \\
    \varphi_1(h\mathbf{L}_n) =& \begin{bmatrix}
        \frac{\sin \gamma}{\gamma}          &         \frac{1-\cos\gamma}{\gamma}          \\
        \frac{\cos\gamma -1 }{\gamma}    &      \frac{\sin \gamma}{\gamma}
    \end{bmatrix} \,, \\
    \varphi_2(h\mathbf{L}_n) =& \begin{bmatrix}
        \frac{1-\cos\gamma}{\gamma^2}  &  \frac{\gamma - \sin \gamma}{\gamma^2}              \\     \frac{\sin \gamma-\gamma}{\gamma^2} & \frac{1-\cos\gamma}{\gamma^2}     
    \end{bmatrix} \,, \\
    \varphi_3(h\mathbf{L}_n) =& \begin{bmatrix}
      \frac{\gamma - \sin \gamma}{\gamma^3}    &      \frac{\cos\gamma -1 +\frac{1}{2}\gamma^2}{\gamma^3}          \\   \frac{ 1 -\frac{1}{2}\gamma^2-\cos\gamma}{\gamma^3}   &   \frac{\gamma - \sin \gamma}{\gamma^3}   
    \end{bmatrix} \,,
\end{align}
where we set $\gamma=h \omega_n^{2p+1}$.

\subsection{Adaptation for 2D PDEs}

In the case of 2D PDEs, it is convenient to write the solution's basis representation as 
\begin{equation}
    u(x, y) = \sum_{m=0}^{M-1} \sum_{n=0}^{N-1}  \hat{u}_{n m} b_m(x)d_n(y),
\end{equation}
where $\hat{\mathbf{u}} \in \mathbb{R}^{M \times N}$ is a 2D tensor.

Again, by choosing as bases $\textbf{b}$ and $\textbf{d}$ some eigenfunctions of the Laplacian, $\mathbf{L}$ becomes separable, turning the application of $\varphi$ into an element-wise operation in $\mathbb{R}^{M \times N}$, which greatly simplifies the necessary calculations.

\end{document}